\documentclass{article}

\usepackage[preprint]{neurips_2025}

\usepackage[utf8]{inputenc} %
\usepackage[T1]{fontenc}    %
\usepackage{hyperref}       %
\usepackage{url}            %
\usepackage{booktabs}       %
\usepackage{amsfonts}       %
\usepackage{nicefrac}       %
\usepackage{microtype}      %
\usepackage{xcolor}         %
\usepackage{array}
\usepackage{graphicx}
\usepackage{cleveref}
\usepackage{comment}
\usepackage[frozencache=true,cachedir=minted-cache]{minted}
\setminted[python]{breaklines}
\usepackage{caption}
\usepackage{subcaption}
\usepackage{wrapfig}
\usepackage{multirow}

\newcommand{\cG}{\mathcal{G}}
\newcommand{\cV}{\mathcal{V}}
\newcommand{\cE}{\mathcal{E}}

\title{Fedivertex: a Graph Dataset based on Decentralized Social Networks for Trustworthy Machine Learning}

\author{%
  Marc Damie\textsuperscript{1,2}\thanks{Corresponding author: \texttt{m.f.d.damie@utwente.nl}}\\
  \And
  Edwige Cyffers\textsuperscript{3}
  \And
  \textsuperscript{1}{\mdseries University of Twente, Enschede, The Netherlands}\\
  \textsuperscript{2}Inria, Villeneuve-d'Ascq, France\\
  \textsuperscript{3}Institute of Science and Technology Austria, Klosterneuburg, Austria\\[0.5em]
}

\begin{document}

\maketitle

\begin{abstract}
Decentralized machine learning -- where each client keeps its own data locally and uses its own computational resources to collaboratively train a model by exchanging peer-to-peer messages -- is increasingly popular, as it enables better scalability and control over the data. A major challenge in this setting is that learning dynamics depend on the topology of the communication graph, which motivates the use of real graph datasets for benchmarking decentralized algorithms. Unfortunately, existing graph datasets are largely limited to for-profit social networks crawled at a fixed point in time and often collected at the user scale, where links are heavily influenced by the platform and its recommendation algorithms.
The Fediverse, which includes several free and open-source decentralized social media platforms such as Mastodon, Misskey, and Lemmy, offers an interesting real-world alternative. We introduce \emph{Fedivertex}, a new dataset of 182 graphs, covering seven social networks from the Fediverse, crawled weekly over 14 weeks. We release the dataset along with a Python package to facilitate its use, and illustrate its utility on several tasks, including a new \emph{defederation} task, which captures a process of link deletion observed on these networks.

\end{abstract}

\section{Introduction}

Decentralized machine learning \cite{hallaji_decentralized_2024} has gained significant popularity in the past years. In this paradigm, each node possesses a local dataset and some computational resources, and collaborates with other participants by exchanging messages through peer-to-peer communication during the training of a global, potentially personalized, machine learning model. In comparison to federated learning \cite{kairouz_advances_2021}, which keeps data local but orchestrates training through a central server, decentralized learning offers additional flexibility as it avoids the bottlenecks and single points of failure that arise from centralized supervision. The shift towards decentralized learning can also be motivated by trust, as the communication graph can reflect users’ chosen collaborations--often referred to as nodes, instances, clients, or participants in this context. The network topology has an impact on the learning dynamics \cite{kavalionak_impact_2021}, particularly in the presence of data heterogeneity \cite{hsieh_non-iid_2020,vogels_relaysum_2021} and in terms of privacy guarantees \cite{mrini2024privacy}.

Decentralized learning has numerous real-world use cases \cite{hallaji_decentralized_2024}, as nodes can represent healthcare institutions or sensors distributed across installations. One of the most compelling applications is for decentralized social networks \cite{bin_zia_toxicity_2022}. In such cases, the graph often captures complex and diverse relationships, which explains its popularity in machine learning. In particular, this motivates various graph learning tasks \cite{tang_graph_2010}, including community detection, node classification, and edge prediction. Social network dynamics also raise interesting questions related to polarization and time-evolving properties of the graph \cite{toivonen_comparative_2009}. Addressing these questions requires access to relevant social network datasets that allow studying these properties.

The Fediverse -- a contraction of ``federation'' and ``universe'' -- provides decentralized and interoperable online social services. It is often seen as an alternative \cite{anderlini_emerging_2022} to major social networks operated by for-profit companies, and it promotes a very different culture. The Fediverse is decentralized across many servers, called \emph{instances}. Anyone can run an instance, which operates independently under the moderation of its owner, and instances collaborate with each other: a user of a given instance can interact with and follow users from other instances. Since 2018, the Fediverse has adopted ActivityPub \cite{activitypub}, a protocol and open standard that provides a client-to-server API for creating and modifying content, as well as a federated server-to-server protocol for delivering notifications and content across servers. This enables interoperability between different instances and software. The diversity of platforms, the growing number of users, and the international impact of the Fediverse make it an interesting object of study for the machine learning community. In particular, agents in the Fediverse tend to be more aware of the potential ethical issues of machine learning than traditional users of social networks \cite{Struett2023}, and more interested in new features and improvements. This aligns closely with the goals of Trustworthy Machine Learning and the paradigm of collaborative learning, where agents are expected to monitor their participation based on expected benefits.

In this work, we provide the first dataset covering multiple software platforms in the Fediverse, called \emph{Fedivertex}, to enable researchers to easily run experiments on decentralized machine learning tasks and to benchmark several graph learning tasks. By surveying seven different platforms and constructing different types of graphs, we are able to capture the diverse dynamics at play in the Fediverse. In particular, a striking difference from many mainstream social networks is the so-called \emph{defederation} process \cite{lai_new_2025}, in which instances choose to sever ties with other instances, often due to a disagreement on moderation or security practices. While new link prediction is often regarded as the primary task for time-evolving graphs \cite{hasan_survey_2011}, a major dynamic in the Fediverse is this complementary edge deletion. Our dataset is the first to enable the study of this phenomenon. The current $14$ distinct timestamps for each graph are a starting point to study the evolution over time, and we plan to continue to update the dataset in the future. More precisely, our contributions are as follows:

\begin{itemize}
\item[(i)] We introduce \emph{Fedivertex}, a large and diverse graph dataset based on the Fediverse. More precisely, our dataset encompasses seven Fediverse platforms, resulting in $182$ graphs: $13$ different graphs each with a sequence of $14$ snapshots obtained through weekly web crawls over a period of three months.
\item[(ii)] We provide a Python package, \texttt{fedivertex}, available through PyPI, to easily access and use our dataset. The package includes built-in preprocessing tools to download and prepare the graphs for machine learning tasks. We demonstrate its usefulness by benchmarking several existing decentralized learning algorithms.
\item[(iii)] We formalize a novel graph analysis task: \emph{defederation prediction}, which aims to predict which edges or nodes will be removed from the graph at the next iteration, and we propose baselines for this task.
\end{itemize}

\section{Related work}
\label{sec:related_works}

\textbf{Decentralized machine learning.}
Federated learning and fully decentralized learning are increasingly studied~\citep{kairouz_advances_2021,Li2020,pmlr-v54-mcmahan17a}, with various algorithms based on gossip~\citep{boyd2006gossip,hendrikx_principled_2022,Hendrikx2020DualFreeSD,pmlr-v119-koloskova20a,dec2017,JMLR:v17:15-292,neglia2020,tang18a} or random walks~\citep{even_stochastic_2023,johansson,hendrikx_principled_2022,Walkman}. These results highlight the importance of communication graphs for the quality of the final model, the speed of convergence in the presence of heterogeneous data~\citep{pmlr-v206-le-bars23a}, personalization \cite{Bellet2018a} and privacy guarantees~\citep{cyffers2024differentially,cyffers2022muffliato,Dekker2023TopologyBasedRP,mrini2024privacy}, which motivates the use of recent real-world social networks.

\textbf{Social network datasets and analysis.}
Machine learning frequently relies on small social networks, such as the Karate Club~\citep{karate} or citation networks~\citep{citeseer,cora}. Several larger digital social networks are also available via platforms like SNAP~\citep{leskovec_snap_2014}, in particular Facebook and Twitter graphs. It has been shown that for-profit platforms influence user graphs, as their recommendations about whom to follow accelerate the triadic closure process and exacerbate inequality in popularity~\citep{effectofrecommendation,zignani_follow_2018}. This motivates the study of social networks that do not follow this trend. In particular, the Fediverse enables analysis at the level of servers rather than at the level of individual users, an approach that captures entities more likely to develop consistent collaboration policies. Prior work on the Fediverse remains limited, often focusing on a single network or on interactions between a fixed pair of networks, and typically does not provide reusable datasets~\citep{agarwal2024decentralised,Hironaka2024,zignani_follow_2018}.

\section{Fedivertex Dataset}

\subsection{Fediverse software and graphs}

In the Fediverse, software is run by servers referred to as  \emph{instances}, without any centralized control or coordination. Each instance hosts a subset of \emph{users} and has its own internal rules and moderation. Despite maintaining sovereignty over their rules and storing data locally, instances are not isolated from each other, as they all use the same protocol and standard: ActivityPub \cite{activitypub}. This protocol enables communication between instances and even across services. For instance, a video from PeerTube can be shared on Mastodon, and the resulting post can be viewed from Misskey -- unlike traditional social network silos, where a Facebook user cannot use their account to read tweets or watch YouTube videos. One can think of this interoperability similarly to email, where a user from one provider (e.g., Gmail) can send a message to another (e.g., Outlook).
ActivityPub includes both a federation protocol -- a server-to-server protocol that allows instances to share information -- and a social API--a client-to-server protocol that allows users to send information to their instance. A user's data is stored on their respective instance but can be duplicated and cached on other instances to be accessible to other users. When two users communicate, only their respective instances -- and possibly a third instance hosting the interaction -- are aware of the message. As a result, data permanence, confidentiality, and moderation depend on the instance.

Fedivertex focuses on interactions between \emph{instances} within a given software platform. In particular, for each social network (except Peertube), the \textbf{federation graph} models the undirected communication graph between instances within that network.
Federation graphs are naturally dense, because two instances are connected with an edge if they have interacted at least once.
We selected seven of the most popular software platforms in the Fediverse to ensure sufficient activity for graph-based analysis. Our selection covers diverse types of social network to reflect a range of communication dynamics. We summarize the dataset in \cref{tab:summary} and present each of them in more detail below.

\begin{table}
\centering
    \caption{Overview of the Fedivertex social networks. The number of instances and users has been extracted on May 13, 2025 from FediDB, a reference database for the Fediverse communities. These numbers are indicative as the networks evolve over time.}
    \label{tab:summary}
\small
\setlength{\tabcolsep}{2pt} %
\renewcommand{\arraystretch}{1.1} %

\begin{tabular}{@{}l@{} *{7}{>{\raggedright\arraybackslash}p{1.78cm}@{}} }
\toprule
 & \multicolumn{1}{l@{}}{Peertube} 
 & \multicolumn{1}{l@{}}{Mastodon} 
 & \multicolumn{1}{l@{}}{Pleroma} 
 & \multicolumn{1}{l@{}}{Misskey} 
 & \multicolumn{1}{l@{}}{Friendica} 
 & \multicolumn{1}{l@{}}{Bookwyrm} 
 & \multicolumn{1}{l@{}}{Lemmy} \\
\midrule
 \textbf{Type} 
 & Video streaming 
 & Micro blogging 
 & Micro blogging 
 & Micro blogging 
 & Micro blogging 
 & Book cataloging 
 & Social news \\
\textbf{1st release}
 & 2018 & 2016 & 2017 & 2014 & 2010 & 2022 & 2019 \\
\textbf{Screenshot}
 & \includegraphics[height=.9cm]{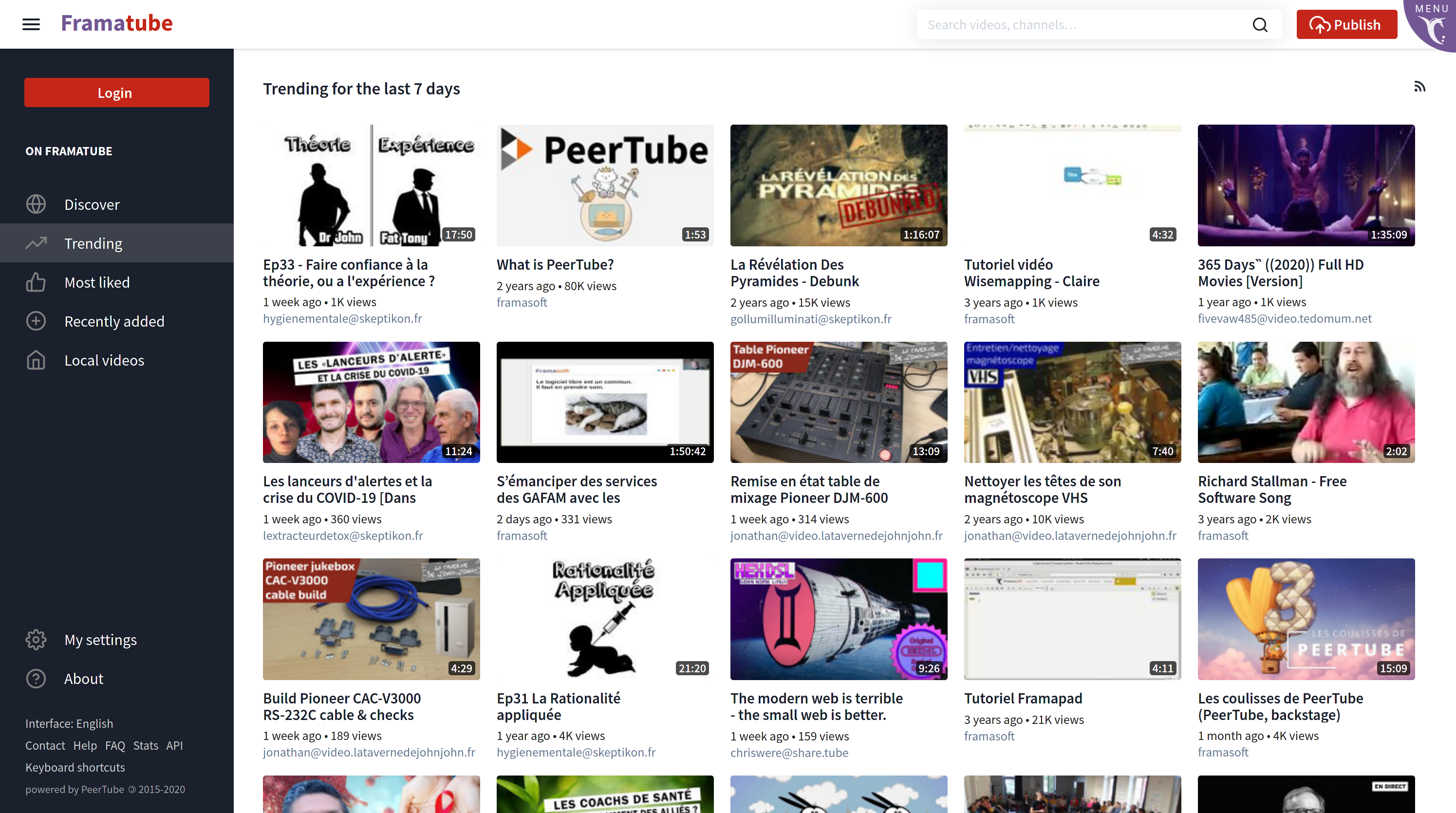}
 & \includegraphics[height=.9cm]{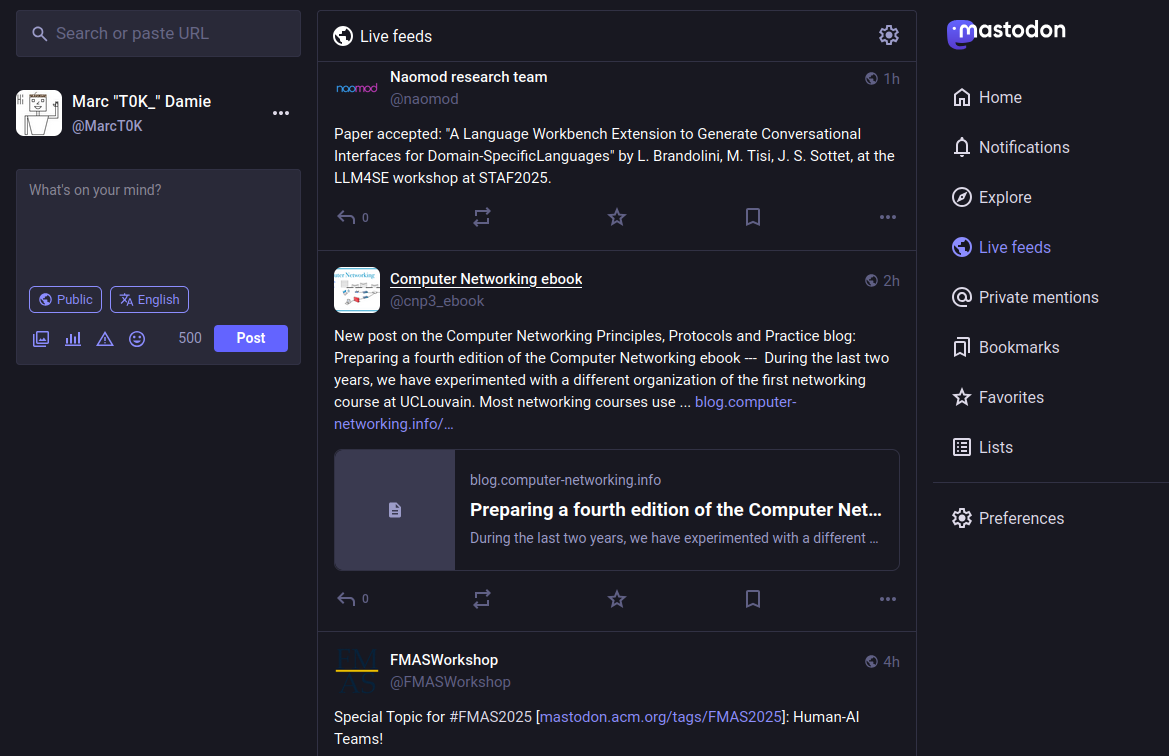}
 & \includegraphics[height=.9cm]{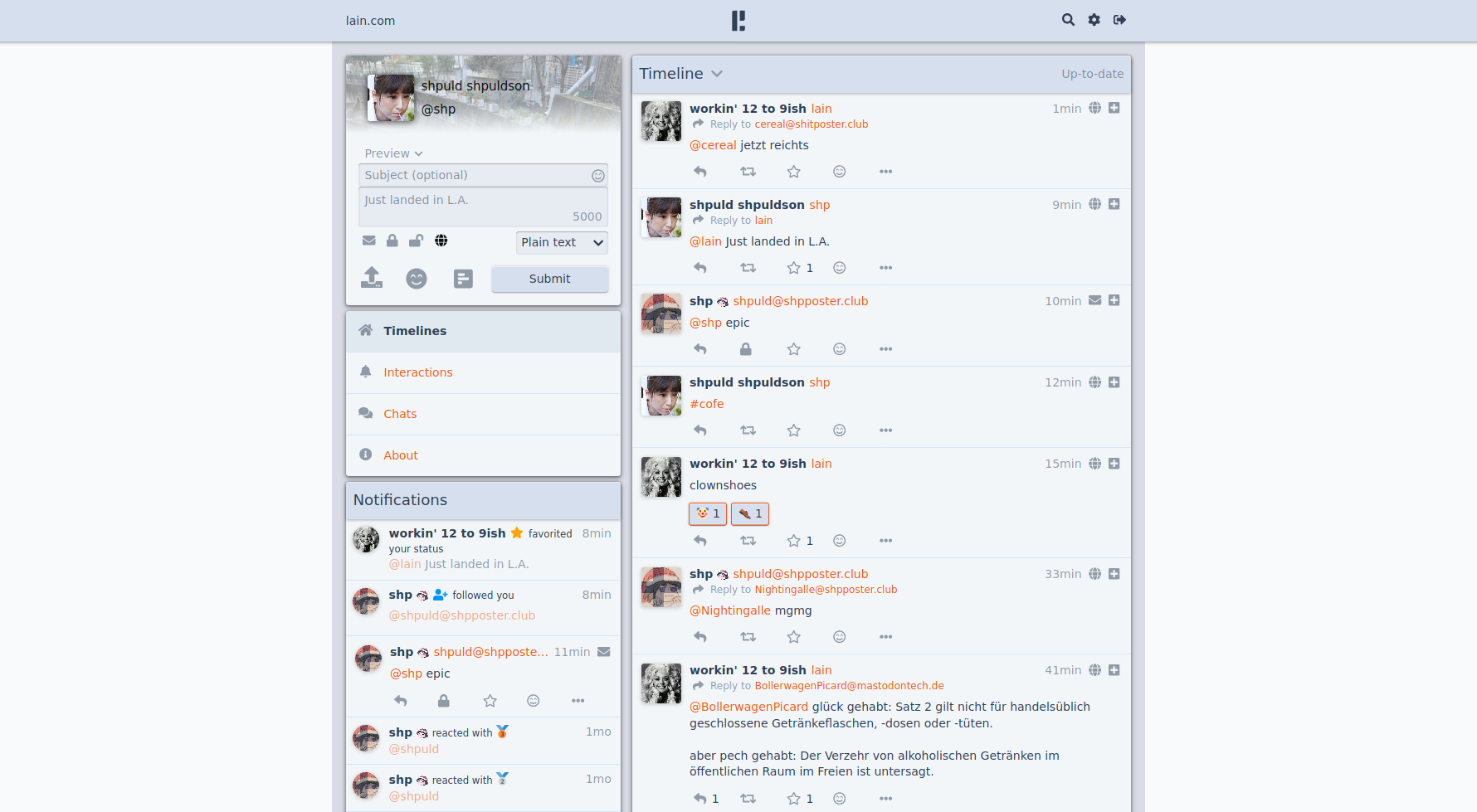}
 & \includegraphics[height=.9cm]{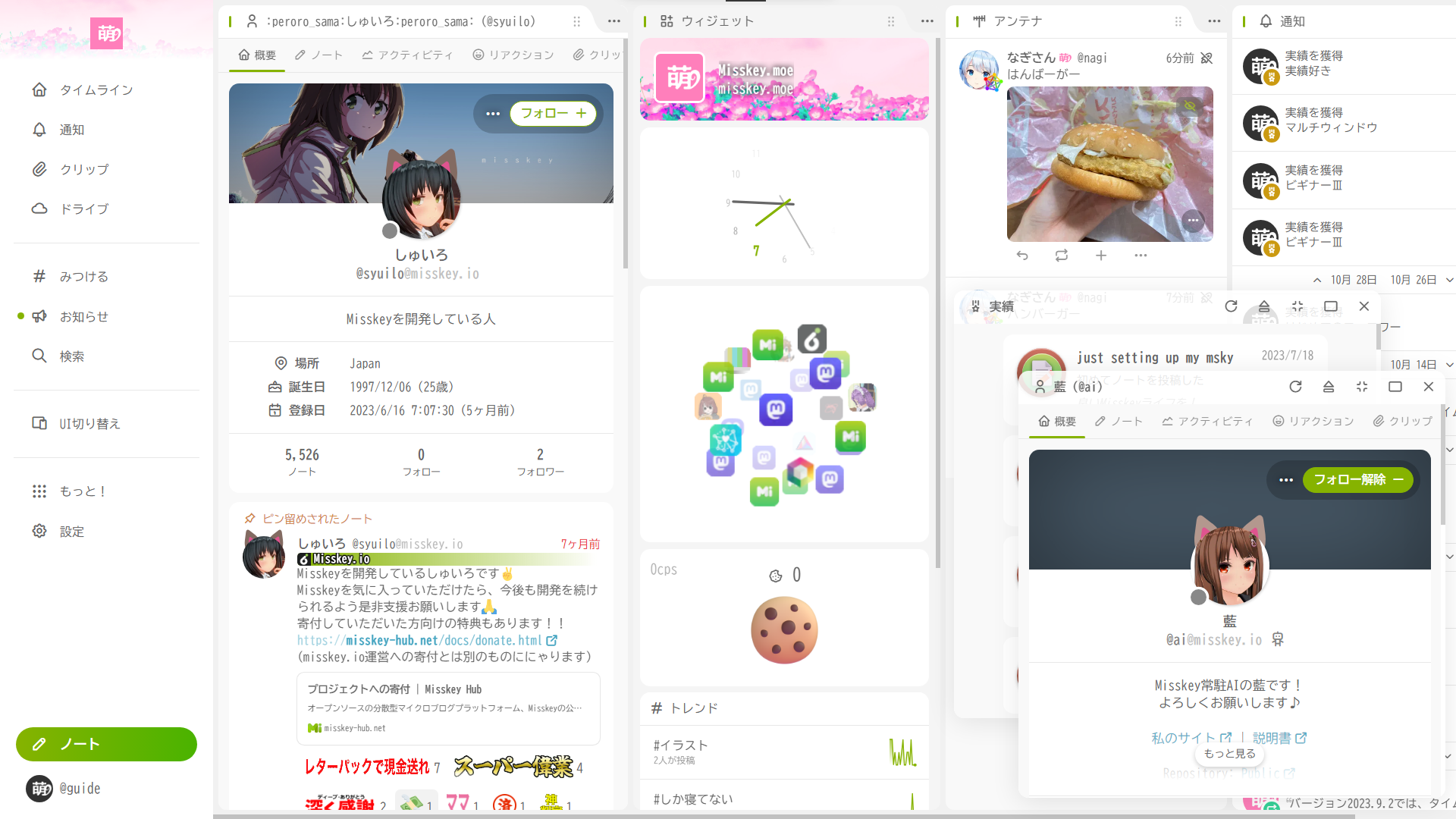}
 & \includegraphics[height=.9cm]{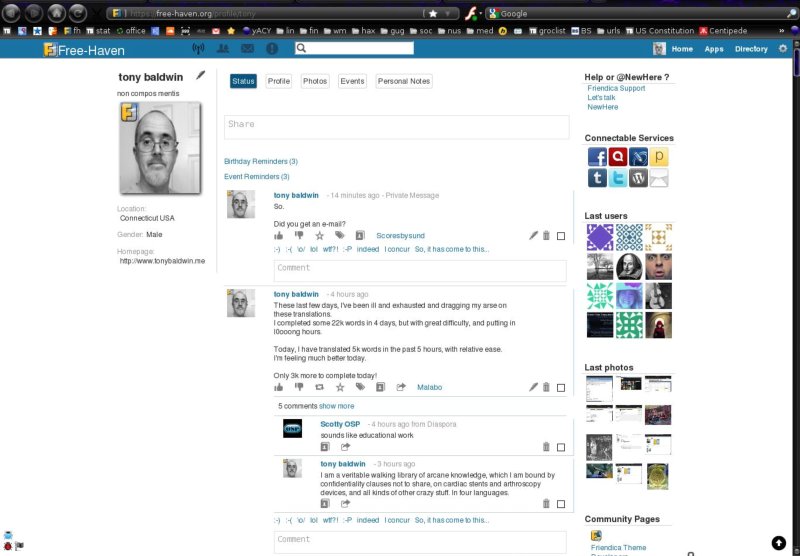}
 & \includegraphics[height=.9cm]{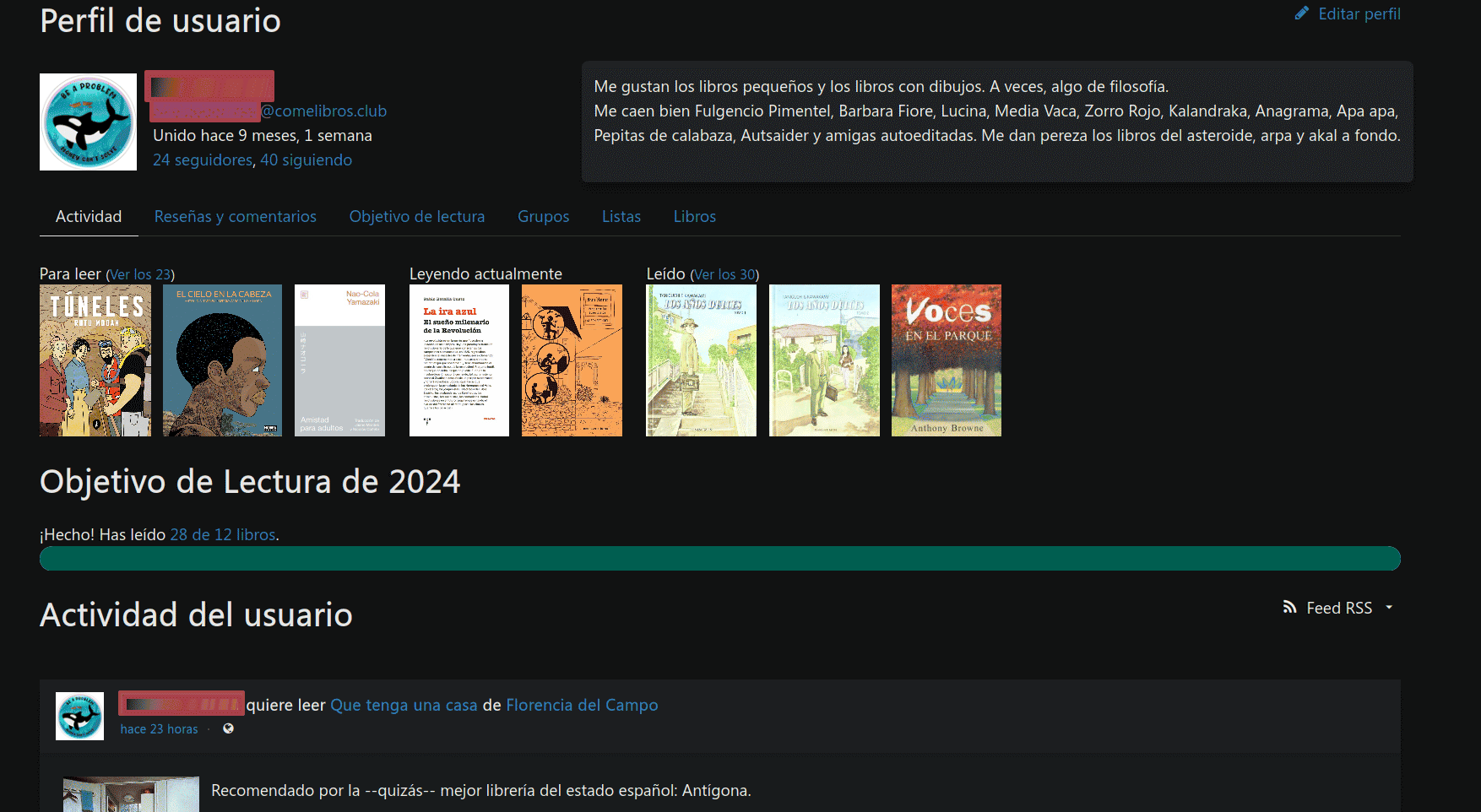}
 & \includegraphics[height=.9cm]{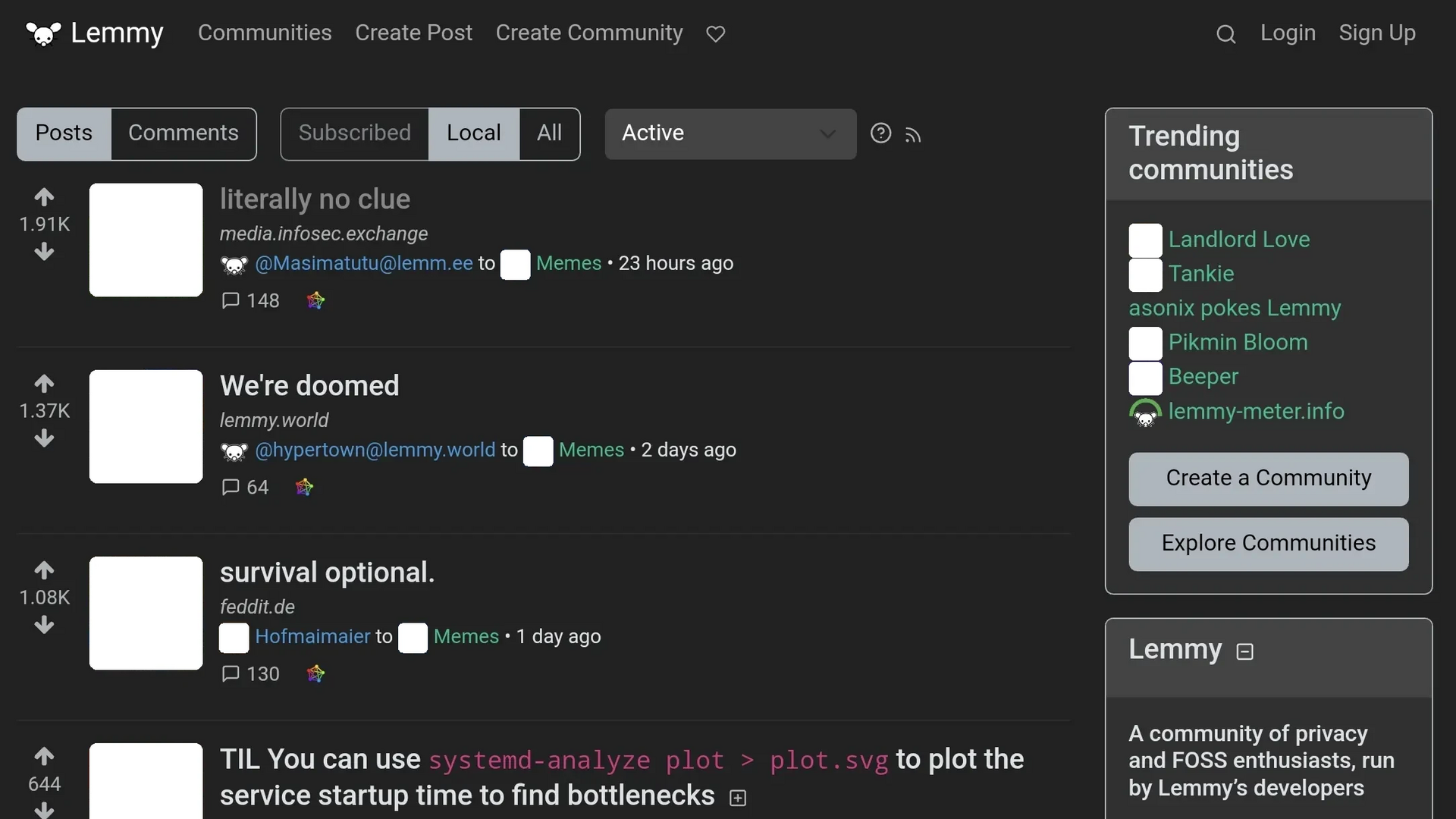} \\
\# \textbf{instances}
 & 1333 & 9652 & 616 & 1206 & 101 & 95 & 564 \\
\# \textbf{users}
 & 583k & 8 102k & 76k & 1 071k & 12k & 51k & 520k \\
\textbf{Graphs}
 & follow
 & federation \hspace{2em}active user
 & federation \hspace{2em}active user
 & federation \hspace{2em}active user
 & federation
 & federation
 & fed. + blocks \hspace{2em}intra-instance\hspace{2em}cross-inst. \\
\bottomrule
\end{tabular}
\end{table}

\subsection{Fediverse social networks}

\textbf{Peertube} provides an alternative to YouTube. Users can watch, bookmark, and comment on videos, subscribe to channels, and create private and public playlists. Video search was added in 2020 with SepiaSearch but remains limited; recommendation features are also a limitation compared to Youtube. An instance $u$ can follow an instance $v$ to let its users see all the videos posted on $v$. We report this \textbf{follow graph} as a directed graph with edges of weight $1$ for following.

\textbf{Mastodon} was created as an alternative to Twitter in 2016 and is supported by the German non-profit organization Mastodon gGmbH. Users post short-form status messages of up to 500 characters, known as ``toots.'' It has experienced several surges in popularity, often in reaction to changes on Twitter, and is sometimes adopted in parallel with it~\cite{Jeong2023ExploringPM}. In addition to the federation graphs, introduced above, we also build a weighted, directed \textbf{active user graph}, with one node per instance. For each instance $u$, we take its $10$k most recently active users. Whenever one of those users follows someone on instance $v$, we increase the edge weight by $1$. Thus weight of the edge from $u$ to $v$ measures how much content seen on $u$ originates from $v$. The graph thus contains self-loop as users follow others on the same instance.

\textbf{Pleroma} is a microblogging software similar to Mastodon, and we thus also report active user graph. Principal difference is allowing longer posts by default, up to $5000$ characters, and offering a lightweight implementation that can potentially run on a Raspberry Pi.

\textbf{Misskey} is a microblogging platform as well, on which we report the active user graph. It was created in 2014 by Japanese software engineer Eiji "syuilo" Shinoda and allows posts of up to $3000$ characters.

\textbf{Friendica} emerged in 2010 as an alternative to Google+ and Facebook, making it the oldest social network in our study, and does not support metadata for active users graph with our crawler.

\textbf{Bookwyrm} allows users to track their reading activity, write book reviews, and follow friends. Launched in 2022 by Mouse Reeve, it can be seen as an alternative to Goodreads.

\textbf{Lemmy} is organized into communities dedicated to specific topics, where users share links and discuss their content. 
Although communities are local to an instance, users can subscribe to those hosted by other instances and participate in discussions across instances. In addition to the federation graph, we report two other graphs. Firstly, the \textbf{intra-instance graph} where the instance $u$ is linked to $v$ if an user of $u$ has published a message on instance $v$. This graph is directed and very sparse. Then, in \textbf{cross-instance graph}, two instances are connected as soon as there exists a pair of users who published a message in the same thread, but possibly on a third instance. This is an undirected graph, denser that the previous one.

\subsection{Fedivertex package and availability}

Our dataset can be directly downloaded from Kaggle \cite{damie_fedivertex_2025}.
To facilitate its use, we also provide a Python package, \texttt{Fedivertex}, which allows users to directly load the graphs in NetworkX format through an easy-to-use interface, as shown in \cref{list:code_example}. 
We facilitate interaction with Fedivertex package by releasing several notebooks to analyze the graphs. Finally, we follow the Gephi convention for graph encoding, allowing the graph CSV files to be opened directly in this software \cite{bastian2009gephi}.

\begin{listing}[tb]
\tiny
\begin{minted}[frame=single,]{Python}
from fedivertex import GraphLoader

loader = GraphLoader()
loader.list_graph_types("mastodon") # List available graphs for a given software, here federation and active_user
G = loader.get_graph(software = "mastodon", graph_type = "active_user", index = 0, only_largest_component = True)
# G contains the Networkx graph of the giant component of the active users graph at the 1st date of collection
\end{minted}
\caption{Code example from the \texttt{Fedivertex} package\label{list:code_example}}
\end{listing}

\subsection{Construction via Web crawling}
For each of the $13$ graphs introduced above, we produce a new version every week (thus presenting 14 different timestamps of each of the graph at the moment of the article submission). To identify all available servers for a given software, we query the Fediverse Observer\footnote{\url{https://fediverse.observer}}, which provides a curated list of Fediverse instances commonly accepted by the community. We then query each of these instances to compute the edges of the graph. Relying on the Fediverse Observer list helps minimize server load and allows us to benefit from existing curation. Notably, the Fediverse Observer's crawler runs daily and is also open-source.
We release the code of our crawler for reproducibility and to allow extensions to other social media or other scraping parameters.

\subsection{Ethical concerns}
\label{subsec:ethical-concerns}
Our work aims to bring more attention to the Fediverse social networks, who could benefit from Trustworthy Machine Learning applications, for instance to assist in moderation task \cite{bin_zia_toxicity_2022} or with user experience and recommendation systems. However, Fediverse software has often been developed to avoid various downsides of hastily deployed machine learning models, from toxicity to invisible filter bubbles, and dark patterns to poor accuracy \cite{biasRS,Matakos2017}. It is thus part of the challenge and the motivation to focus on tasks where the improvement for the user overcomes the possible drawbacks. Hence, we decide to illustrate our datasets only on tasks that could respect the Fediverse mindset. 

The collection of the dataset raises two major problems, the privacy and the possible disturbance for the service. Concerning the impact of the crawling, we designed our crawler to minimize the impact on the servers, by limiting the size of the queries, using a delay of $0.4$ second between requests on a given instance to avoid disturbing their operations. We do not disguise our requests into real users' ones and we use a clearly identifiable user agent providing a direct contact to us. We respect the policy of the instances by following the instructions given by \texttt{robots.txt} files.  

In order to respect the privacy of the users, we did several design choices. First, our dataset is instance-based and not user-based, which is a better granularity for privacy. Second, we only report general metadata but never store actual messages or content from the social networks. Third, we only use public API endpoints, which do not require accounts on this platform: we do not try to circumvent these privacy practices by creating accounts to access more information. Forth, we also respect informal privacy practices. For instance, we ignore all users using the hashtag \#NOBOT in their profile as it is an informal anti-bot policy on Mastodon. Finally, we limit the access by post-processing instances names to avoid direct clicking links. The whole scraping process was supervised by the legal department of our institutes to ensure compliance with GDPR.

\section{Dataset analysis}
\label{sec:dataset-analysis}
\subsection{Dataset properties}

\begin{figure}[thb]
     \centering
     \begin{subfigure}[b]{0.43\textwidth}
         \centering
         \includegraphics[width=.9\textwidth]{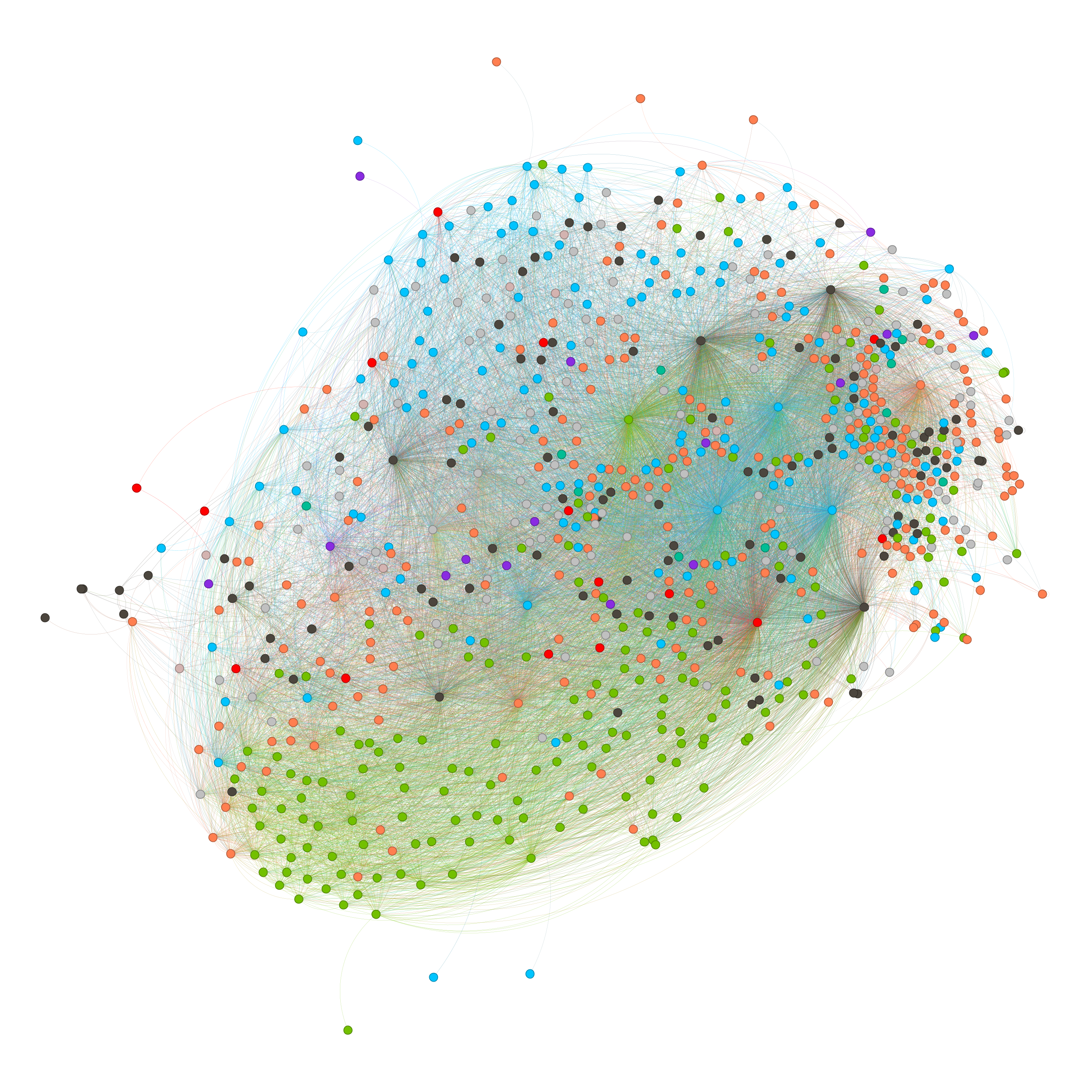}
         \caption{Peertube Follow graph on April 28th 2025. Colors encode the official language of each instance, with green for French, blue for English, black for German, red for Italian and orange when there is no official language. Interactive version: \url{https://marc.damie.eu/peertube_graph/index.html}}
         \label{fig:peertube}
     \end{subfigure}
     \hfill
     \begin{subfigure}[b]{0.5\textwidth}
         \centering
         \includegraphics[width=.9\textwidth]{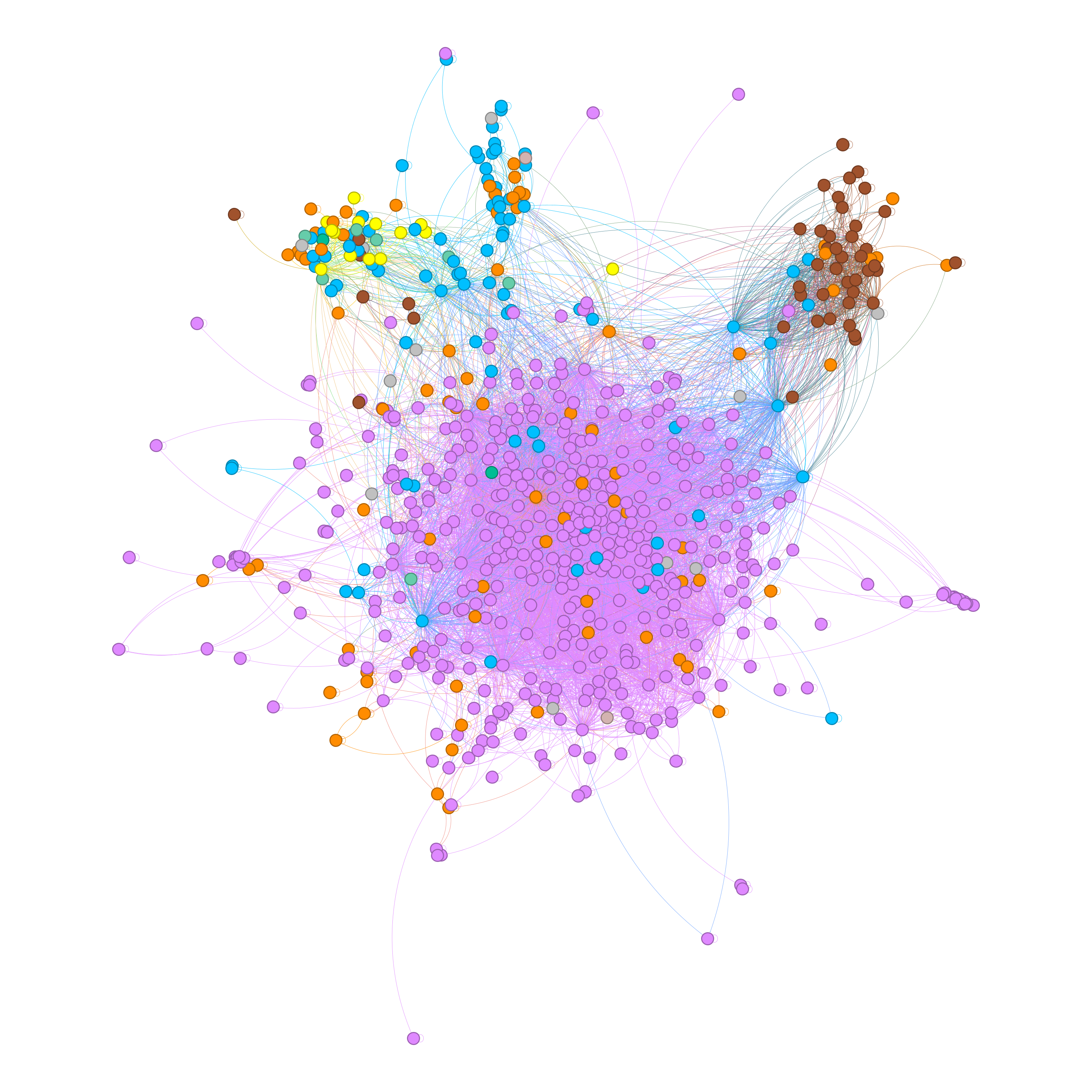}
         \caption{Misskey Active User graph after removal of \texttt{Misskey.io} on May 7th, same colors than left figure, and Chinese in yellow, Japanese in purple and Korean in brown. Interactive version: \url{https://marc.damie.eu/misskey_graph/index.html}}
         \label{fig:misskey}
     \end{subfigure}
     \caption{Examples of graphs communities based on the official languages in Fedivertex dataset}
     \label{fig:language}
\end{figure}

The Fedivertex dataset presents different characteristics depending on both the graph and the software considered. We share in this section a few observations. First, all graphs are provided with their temporal evolution, with new nodes and edges appearing or disappearing each week. For readability, we present only a subset of the graphs and refer the reader to \cref{app:fig} and the notebooks for a more systematic review\footnote{\url{https://www.kaggle.com/code/marcdamie/exploratory-graph-data-analysis-of-fedivertex}}. 

\textbf{Communities.} Fedivertex contains language labels for Peertube, Lemmy, Bookwyrm, and Misskey. For Peertube, the labels are directly extracted from the instance descriptions. For the others, we infer the label from the language of the instance description. The labels can be easily used for label prediction tasks through our API, as described in \cref{lst:language_extraction}. We report on \cref{fig:peertube} the labels for the Peertube graph, which exhibits a clear French-speaking community and Misskey, which is dominated by the Japanese community but also exhibits smaller Korean and Chinese communities, and we refer to \cref{sec:communities} for the associated prediction task. 

\begin{listing}[tb]
\tiny
\begin{minted}[frame=single,]{Python}
G = loader.get_graph(software = "misskey", graph_type = "active_user")
lang_labels = [data["description_language"] for node_name, data in G.nodes(data=True)]
\end{minted}
\caption{Code example to extract language information from the Misskey active user graphs}
\label{lst:language_extraction}
\end{listing}

\begin{table}[t]
\centering
\caption{Small world properties of the Fedivertex graphs and SNAP Social graphs}
\label{tab:fediverse-properties}
\begin{tabular}{lc>{\centering}p{1.2cm}>{\centering}p{1.5cm}>{\centering}p{1.5cm}c}
\toprule
\textbf{Graph} & \textbf{Directed} & \textbf{Avg. Degree} & \textbf{Avg. Path Length} & \textbf{Cluster. Coef.} & \textbf{Small-world $\sigma$} \\
\midrule
Bookwyrm Federation     & No    & 55  & 1.23 & 0.89 & 1.14 \\
Friendica Federation    & No    & 156 & 1.41 & 0.85 & 1.41 \\
Lemmy Federation        & No    & 355 & 1.18 & 0.94 & 1.15 \\
Lemmy Intra-instance    & Yes   & 13  & 2.03 & 0.69 & 3.80 \\
Lemmy Cross-instance    & Yes   & 42  & 1.60 & 0.82 & 1.89 \\
Mastodon Active user 	& Yes   & 125 & 2.09 & 0.73 & 21.95 \\
Misskey Federation      & No    & 317 & 1.66 & 0.76 & 2.21 \\
Misskey Active user 	& Yes   & 19  & 2.36 & 0.62 & 20.78 \\
Peertube Follow         & Yes   & 23  & 2.82 & 0.53 & 4.45 \\
Pleroma Federation      & No    & 269 & 1.64 & 0.82 & 2.26 \\
Pleroma Active user 	& Yes   & 7   & 3.95 & 0.30 & 2.53 \\
\midrule
\midrule
Facebook Ego            & No    & 47  & 3.7  & 0.61 & 39.44 \\
Github                  & No    & 29  & 3.25 & 0.14 & 31.49 \\
Wikipedia Vote          & No    & 15  & 3.25 & 0.17 & 519.64 \\
\bottomrule
\end{tabular}
\end{table}
\textbf{Graph statistics.} We report few metrics in \cref{tab:fediverse-properties} that are usually applied for social networks. All the reported graphs exhibit small-world properties to an extend, as they satisfies $\sigma>1$, which means that a node is more likely to connect to the neighbors of its neighbors and the average path length is small. However, the strength of the small world properties depends on the software.

\begin{figure}
    \centering
    \captionsetup{format=plain}
 \includegraphics[width=0.5\textwidth]{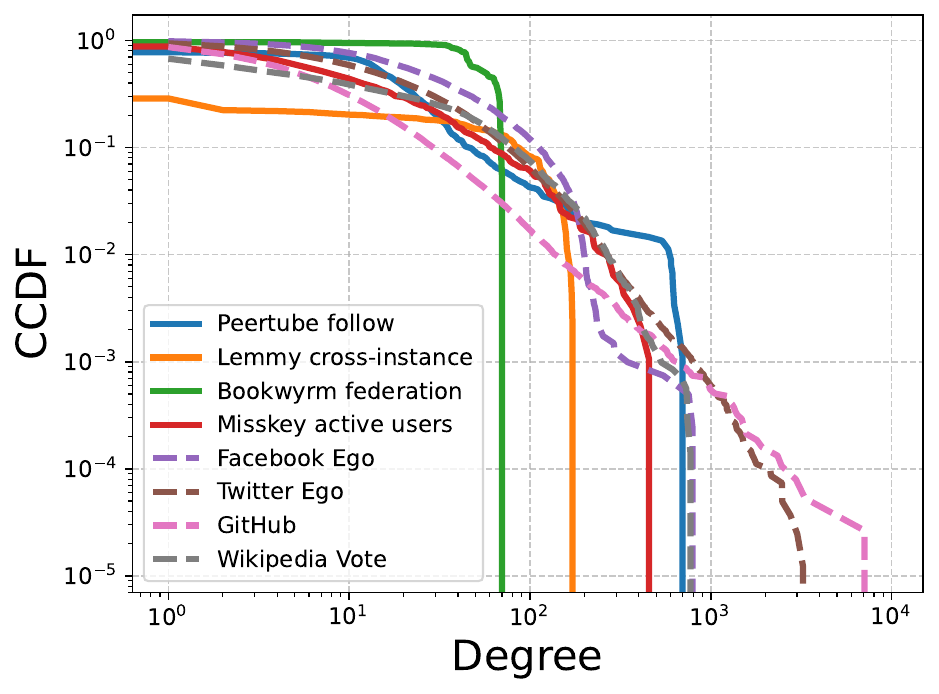}
    \caption{Complementary Cumulative Distribution Function (CCDF) of the degree for several Fedivertex graphs and other widely-used social networks. A normalized version is available in \cref{app:code}}
    \label{fig:comparison}
    \vspace{-12pt}
    \end{figure}

\subsection{Comparison with existing graph datasets}
We compare our graph with the most popular social network graphs from SNAP~\cite{leskovec_snap_2014}. We include the Wikipedia Vote graph~\cite{leskovec_predicting_2010}-- which encodes all Wikipedia voting data up to January 2008, representing each user who participated in a vote as a node and adding a directed edge from node~$i$ to node~$j$ if user~$i$ voted for user~$j$-- as well as the Twitter and Facebook Ego graphs~\cite{leskovec_learning_2012}, corresponding respectively to the Follow and Friend relationships. We also include the GitHub graph~\cite{rozemberczki2019multiscale}, where nodes are developers who have starred at least 10 repositories and edges represent mutual follower relationships.
On \cref{fig:comparison}, while GitHub and Twitter exhibit the classical power-law decay over much of the support of the distribution, consistently with preferential attachment networks~\cite{barabasi2016network}, the degree distribution of Fedivertex is more diverse. We note some similarities between the Facebook Ego graph and the Peertube follow graph, with a smooth concave decay followed by a short power-law tail for the most popular users. Most of the graphs, however, exhibit only the a concave curve, which suggests that the attachment dynamics in Fedivertex differ from those in traditional social networks, allowing for a smoother distribution of node importance. This difference with Fedivertex is confirmed by the statistics of \cref{tab:fediverse-properties}: Fedivertex has a wider range of average path length (from $1.18$ to $3.95$ versus from $3.25$ to $3.7$) and of degree (from $7$ to $355$ versus $15$ to $47$) and have smaller small-world $\sigma$.
A more systematic comparison is available in this notebook\footnote{\url{https://www.kaggle.com/code/marcdamie/fedivertex-vs-snap-social-graphs/notebook}}.

\section{Applications}
\subsection{Decentralized machine learning and statistics}
Fedivertex graphs are particularly well adapted to experiments testing fully decentralized machine learning, as they provide a credible use-case scenario. We illustrate this by reproducing the main figures of \cite{cyffers2024differentially}. This paper proposes training a global model with differential privacy by performing a random walk on the communication graph: at each step, the stochastic gradient is computed on the local dataset of the current node and sanitized by adding Gaussian noise. The paper derives privacy guarantees in the Pairwise Network Differential Privacy setting, where each pair of nodes has a specific privacy budget depending on their relative position, a high budget corresponds to a greater risk of leaking information. In particular, the paper establishes a connection between the structure of these privacy budgets and the communicability of the graph, showing that nodes close to each other have higher privacy budget than far apart ones. Using graphs with different topologies is interesting to verify that similar patterns appear for privacy losses and known graph quantities such as centrality or communicability. Finally, the paper claims to be quite efficient in terms of privacy–utility trade-offs.  

\begin{figure}[thb]
     \centering
     \begin{subfigure}[b]{0.32\textwidth}
         \centering
         \includegraphics[width=\textwidth]{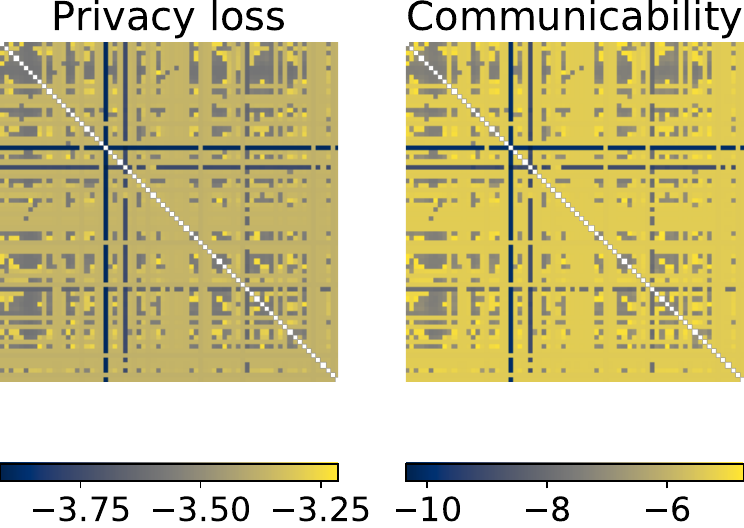}
         \caption{Bookwyrm federation graph communicability and privacy loss,
logarithmic scale}
         \label{fig:dprwbookwyrm}
     \end{subfigure}
     \hfill
     \begin{subfigure}[b]{0.32\textwidth}
         \centering
         \includegraphics[width=\textwidth]{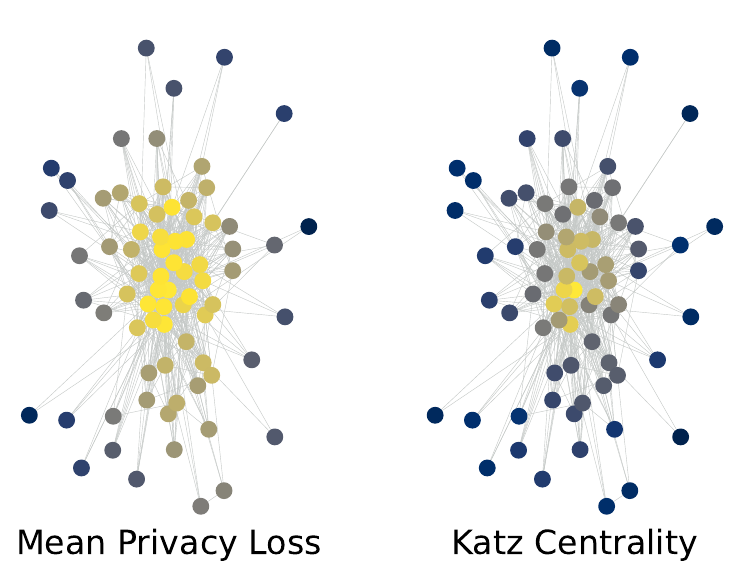}
         \caption{Mean privacy loss and Katz centrality for Lemmy intra-instance graph}
         \label{fig:dprwlemmy}
     \end{subfigure}
     \hfill
     \begin{subfigure}[b]{0.32\textwidth}
         \centering
         \includegraphics[width=\textwidth]{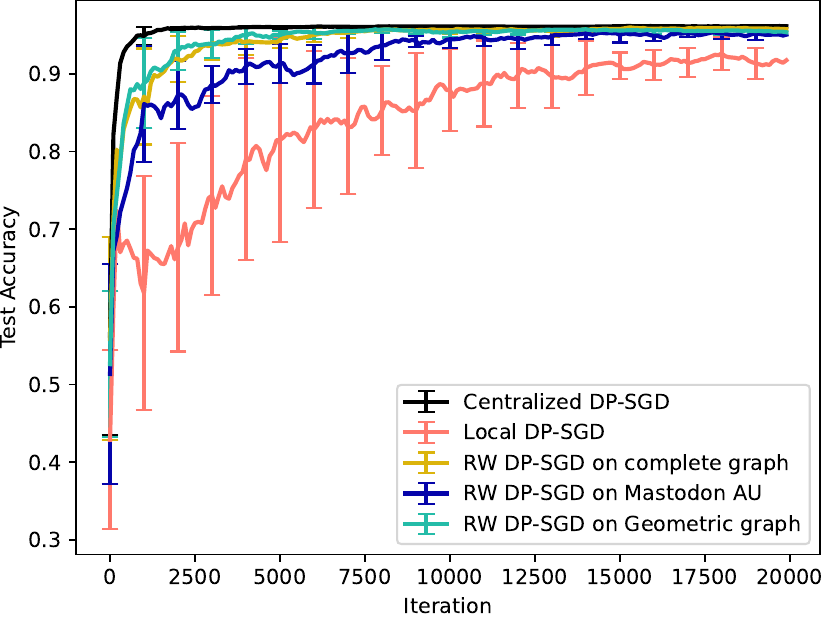}
         \caption{Test accuracy on private logistic regression on the Houses dataset for graph with $3852$ nodes}
         \label{fig:masto}
     \end{subfigure}
     \caption{Numerical experiments reproducing the results of \cite{cyffers2024differentially} with Fedivertex graphs}
     \label{fig:dprw}
\end{figure}

On \cref{fig:dprw}, we see that the link between communicability and privacy budgets is clear on Fedivertex graphs, with the same patterns visible in \cref{fig:dprwbookwyrm} and \cref{fig:dprwlemmy}. However, it also shows that real-world graphs can be more challenging in terms of convergence, as \cref{fig:masto} exhibits slower convergence on the Mastodon active-user graph than on a synthetic graph with the same number of nodes. This could be explained by the presence of nodes with low centrality, typically connected by only a single edge, which makes it harder for the random walk to visit them frequently enough within the chosen number of steps compared with the more regular graph tested in the original paper. To ease comparison, we provide more background on the task and the original figures in \cref{app:code}.

\subsection{New links and defederation prediction}
\label{sec:defederation}

A key feature of Fedivertex is its support for temporal graph analysis, as we release weekly snapshots for each graph. Understanding temporal changes in graphs is seen as a major challenge in graph learning \cite{Majhi2022, Rossi2020TemporalGN}. This is particularly relevant in social networks, where the creation of links often speeds up the triadic closure of the graph, as friends of friends tend to become friends over time \cite{MFforlinkpred, zignani_follow_2018}, especially when platforms actively recommend new connections \cite{effectofrecommendation}. We refer to \cref{tab:fediverse-properties} for Fedivertex graphs' clustering coefficients.

\begin{figure}[bt]
     \centering
     \begin{subfigure}[b]{0.32\textwidth}
         \centering
         \includegraphics[width=\textwidth]{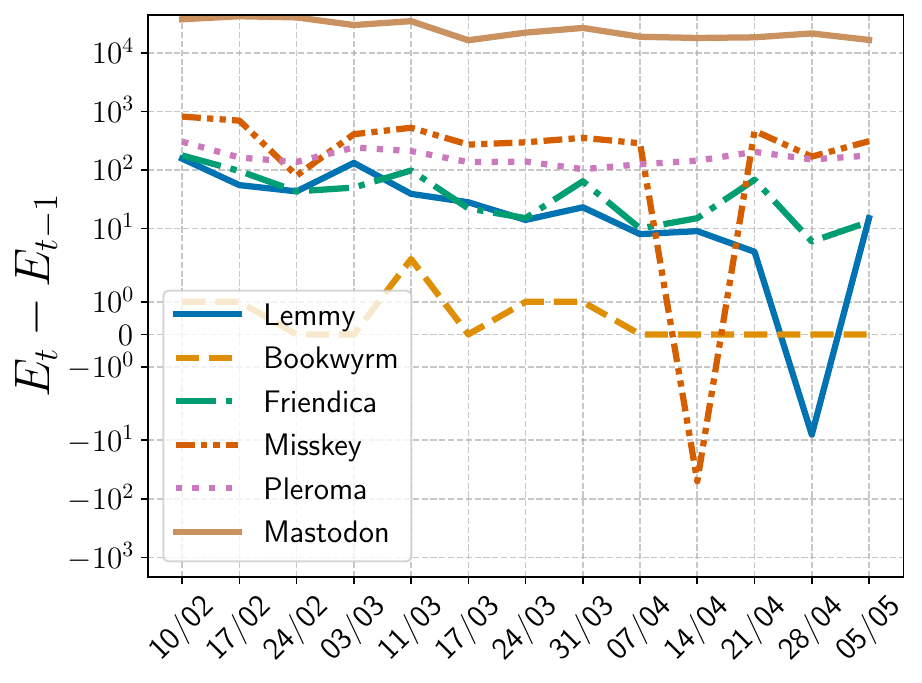}
         \caption{Variation  of the number of edges over time, on the subgraph of the nodes appearing at all iterations for federation graphs, logarithmic scale}
         \label{fig:evofed}
     \end{subfigure}
     \hfill
     \begin{subfigure}[b]{0.32\textwidth}
         \centering
         \includegraphics[width=\textwidth]{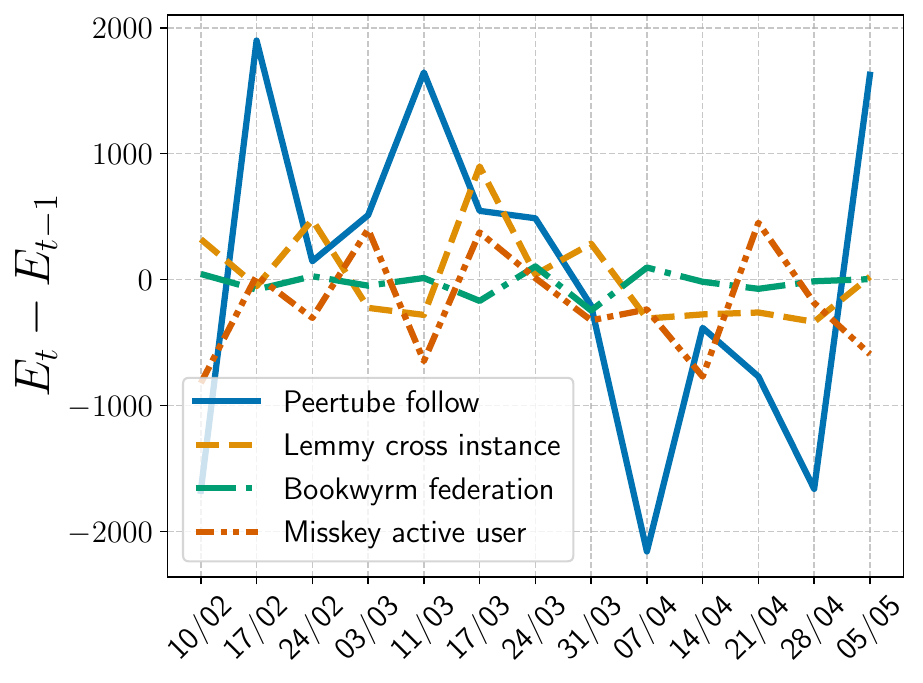}
         \caption{Variation  of the number of edges over time, on the subgraph of the nodes appearing at all iterations, for various type of graphs}
         \label{fig:evoedges}
     \end{subfigure}
     \hfill
     \begin{subfigure}[b]{0.32\textwidth}
         \centering
         \includegraphics[width=\textwidth]{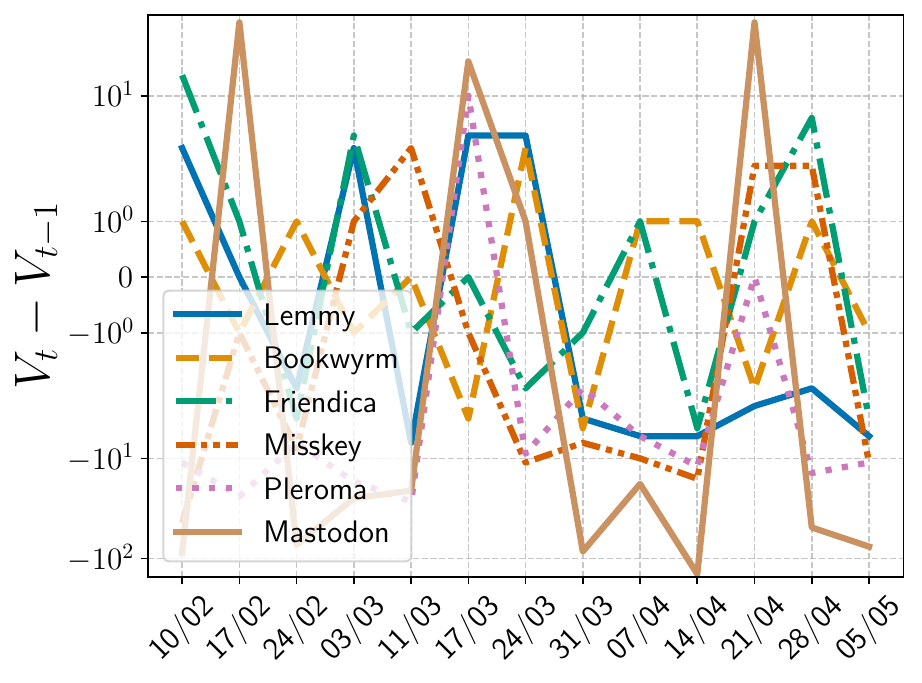}
         \caption{Variation of the number of nodes over time for the six software of Fedivertex, logarithmic scale}
         \vspace{10pt}
         \label{fig:evonode}
     \end{subfigure}
     \vspace{-3pt}
     \caption{Temporal evolution of Fedivertex graphs}
     \label{fig:temp}
\end{figure}

An interesting behavior observed in the Fediverse is that edges between instances are sometimes deleted, a phenomenon that has received little attention so far, likely because edge deletions are uncommon in other social networks. However, in the context of Fedivertex, predicting deletions is interesting for several reasons. First, in some platforms, deletions are as important and can dominate the change in the number of edges and nodes, as seen in \cref{fig:temp}. Secondly, avoiding centralization around a single central server is a key challenge in the Fediverse, and understanding defederation \cite{lai_new_2025} could help maintain a sufficiently decentralized structure. Finally, new link prediction and edge deletion are complementary tasks that may benefit from being studied jointly.

From \cref{fig:temp}, we observe that federation graphs are the most stable over time, as one could expect from their construction in comparison to active users or cross-instance graphs, where activity can fluctuate. However, while some networks grow during the studied period (Friendica federation and Lemmy cross-instance), others show variations dominated by the loss of edges (such as the Pleroma federation or Misskey active users). More precisely, by restricting our analysis to the subgraph of the nodes present at all $14$ dates, most federation graphs have an increasing number of edges, with sometimes sharp drops as we can see for Misskey and Lemmy in \cref{fig:evofed}. Overall, the network is thus growing, but also shows defederation peaks that are quicker than our weekly crawling. In \cref{fig:evoedges}, we can see that the other graphs do not share this clear increasing trend, but tend to alternate between more edge creation and more edge deletion. Finally, the number of nodes itself varies over time, with new servers appearing and others being deleted, leading to the complex evolution reported in \cref{fig:evonode}.

We formally introduce the defederation prediction task. Let $\cG_t = (\cV_t, \cE_t)$ be the graph collected at time $t$ and compare it to the graph $\cG_{t'} = (\cV_{t'}, \cE_{t'})$ collected at $t' > t$. Let $\cG_c$ be the subgraph induced by $\cV_c = \cV_t \cap \cV_{t'}$; the goal is to predict all edges either created $(e \in (\cV_c \times \cV_c) \cap (\cE_{t'} \setminus \cE_t))$ or deleted $(e \in (\cV_c \times \cV_c) \cap (\cE_t \setminus \cE_{t'}))$ between $t$ and $t'$. The possible new edges lie in the set $(\cV_c \times \cV_c) \setminus \cE_t$, whereas the deleted ones are in $\cE_t$, a set significantly smaller if the graph is sparse.
Similarly, one could predict nodes that drop from the graph. In particular, reliable prediction could help detect instances which stopped running because of technical problems despite being active in the graph, and possibly provide technical help to such instances. More formally, the goal would be to predict $\cV_t \setminus \cV_{t+1}$ given $\cG_t$.

\begin{table}[tb]
    \centering
    \caption{Comparison of Adamic-Adar (AA), Common Neighbors (CN), Jaccard and Random method to predict new edge (Add) and edge deletion (Del) on different graphs by reporting the number of correct predictions in Top-$K$ scores (the higher the better). We report the average of three runs over disjoint periods of time.}
    \label{tab:defe}
    \begin{tabular}{lcccccccc}
      \toprule
      \textbf{Graph} 
        & \multicolumn{2}{c}{\textbf{AA}} 
        & \multicolumn{2}{c}{\textbf{CN}} 
        & \multicolumn{2}{c}{\textbf{Jaccard}}  
        & \multicolumn{2}{c}{\textbf{Random}} \\
      \cmidrule(lr){2-3} \cmidrule(lr){4-5} \cmidrule(lr){6-7} \cmidrule(lr){8-9}
        & Add & Del 
        & Add & Del 
        & Add & Del 
        & Add & Del \\
      \midrule
      Mastodon AU (Top $1000$)  & $38$ & $10$ & $40$ & $9$ & $14$ & $10$ & $0.7 \pm 0.5$ & $6 \pm 1.4$ \\
      Misskey AU (Top $200$)   & $4.3$ & $1.3$ & $4.3$ & $1.7$ & $0.7$ & $2$ & $0.2 \pm 0.2$ & $2 \pm 0.8$ \\
      Misskey federation (Top $1000$)  & $494$ & $62$ & $491$ & $63$ & $396$ & $60$ & $42 \pm 6$ & $51 \pm 3.7$ \\
      \bottomrule
    \end{tabular}
  \end{table}

New link prediction can be done based on the topology \cite{LibenNowell2007}, by using the fact that similar nodes are more likely to connect. Methods are thus often based on computing scores for each possible pair of nodes, and then return as prediction the edges with the highest scores. Common scores include the number of common neighbors, the Jaccard score, and the Adamic-Adar score. These scores are then evaluated by looking among the top-$K$ predictions how many are indeed new edges, as we report in \cref{tab:defe}. 
Intuitively, deletion could be seen as the opposite of edge creation, so we propose as a baseline, to return the edges with the lowest scores. However, this approach has limited success for federation graphs. We believe this might be due to defederation being extremely quick, and thus the granularity of our current dataset does not seem sufficient to achieve better than random. An interesting future work could be to use our crawler with higher frequency during defederation periods. It might also indicate that other methods should be developed for this task, opening interesting questions for future work. We refer the reader to \cref{app:defederation} for more analysis.

\subsection{Community detection}
\label{sec:communities}

The Fedivertex social networks are used by many communities that might overlap. The same communities might also use several of the Fedivertex software platforms. To illustrate the feasibility of community detection with our dataset, we use the official languages of each instance as ground truth labels, as shown in \cref{fig:language}, and test three algorithms: Louvain \cite{blondel_fast_2008}, Greedy Modularity \cite{clauset_finding_2004}, and Label Propagation \cite{cordasco_community_2010} on the Peertube follow graph and the Misskey active user graph. To avoid many unique labels, we keep only the $5$ most represented languages in each graph and ignore the other nodes using other languages. Results are in \cref{tab:community-detection}.
We assess the quality of this detection using the Adjusted Rand Index (ARI) and the Adjusted Mutual Information (AMI). The Rand index corresponds to the proportion of node pairs belonging to the same cluster that are classified as such (i.e., the sum of true positives for all classes) over all node pairs, and the adjusted version corresponds to normalization with respect to a random clustering. Similarly, the AMI score corresponds to the mutual information between the ground truth and predicted labels, adjusted for chance.
Finally, we report modularity for each clustering -- a metric for unsupervised clustering that assesses the inherent quality of the partitioning. This suggests that other labels might be suitable as well for clustering the graphs.
No method dominates in this benchmark, highlighting that our graphs exhibit diverse structures which may challenge algorithms in different ways. Experiments are in notebook.\footnote{\url{https://www.kaggle.com/code/marcdamie/community-detection-on-fedivertex}}

Other Fedivertex graphs can be used for community detection, and additional labels could be derived from the data, for example, based on the names of the instances or their official descriptions. It is also possible to track the evolution of communities over time.

\begin{table}[th]
    \centering
    \caption{Performance of several community detection algorithms (average of 100 runs for Louvain).}
    \begin{tabular}{llccc}
    \toprule
         \textbf{Graph} & \textbf{Algorithm} &  \textbf{ARI} & \textbf{AMI} & \textbf{Modularity}\\\midrule
         \multirow{3}{*}{Peertube follow} &  Louvain & 0.055 & \textbf{0.123} & \textbf{0.2168} \\
          & Greedy Modularity & \textbf{0.061} & 0.110  & 0.209 \\
          & Label Propagation & 0.008 & 0.029  &  0.003 \\
          \midrule
          \multirow{3}{*}{Misskey active user} &  Louvain & 0.097 & 0.250 & 0.611 \\
          &  Greedy Modularity & 0.014 & 0.165  & \textbf{0.513} \\
          &  Label Propagation & \textbf{0.229} & \textbf{0.255}  & 0.027 \\
          \bottomrule
    \end{tabular}
    \label{tab:community-detection}
\end{table}
\section{Conclusion}

In this work, we introduce \texttt{Fedivertex}, a dataset modeling interactions between instances across several software platforms of the Fediverse. This is the first dataset publicly released to enable reproducible experiments on graphs from the Fediverse, and it allows the study of more diverse graph dynamics than existing social network datasets. We hope that these graphs can foster machine learning research on this topic and contribute to the development of trustworthy decentralized machine learning, notably on the Fediverse. Among possible applications, this dataset could support the development of decentralized spam detection, the prediction of new or deleted links, the prevention of instance shutdowns through early prediction, and many other tasks related to time-evolving graphs.

\begin{ack}
    This work has benefited from French State aid managed by the Agence Nationale de la Recherche (ANR) under France 2030 program with the reference ANR-23-PEIA-005 (REDEEM project) and was supported by the Netherlands Organization for Scientific Research (NWO) in the context of the SHARE project [CS.011]. This work was supported in part by the Austrian Science Fund (FWF) [10.55776/COE12]. Authors thank the Inria Magnet team, without whom this work would not have been possible.
\end{ack}

\bibliographystyle{plain} %
\bibliography{refc.bib}

\appendix
\section{Further analysis of Fedivertex}
\label{app:fig}

This appendix presents several additional plots complementary to the figures presented in the main text.
In particular, we propose to further analyze the degree distribution of our graphs, and the language distribution within different Fediverse social media.

\begin{figure}[htb]
  \centering
  \begin{subfigure}[b]{0.24\textwidth}
    \centering
    \includegraphics[width=\linewidth]{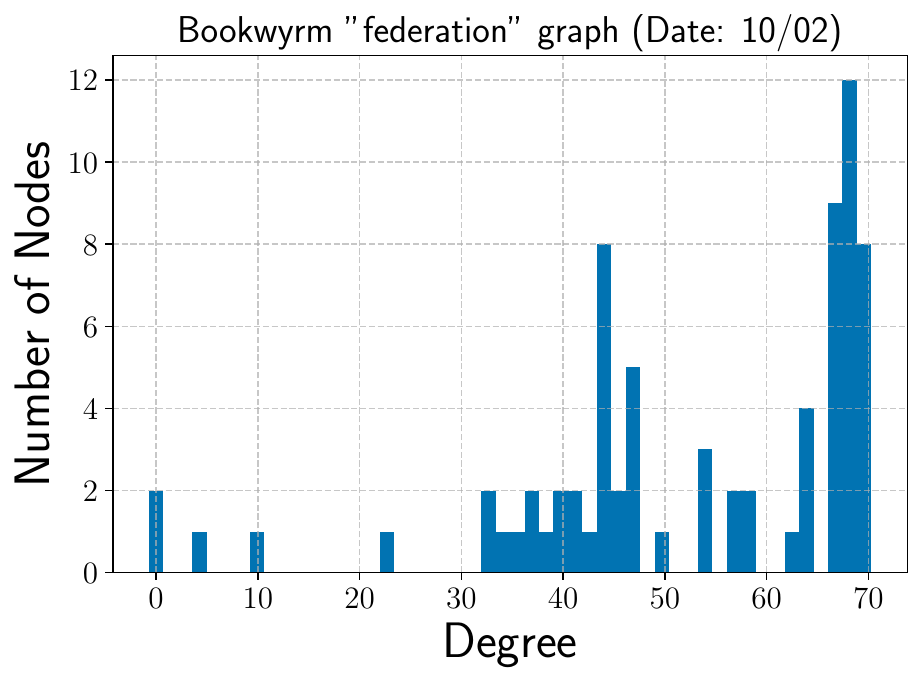}
    \caption{BookWyrm}
    \label{fig:hist_bookwyrm_fed}
  \end{subfigure}\hfill
  \begin{subfigure}[b]{0.24\textwidth}
    \centering
    \includegraphics[width=\linewidth]{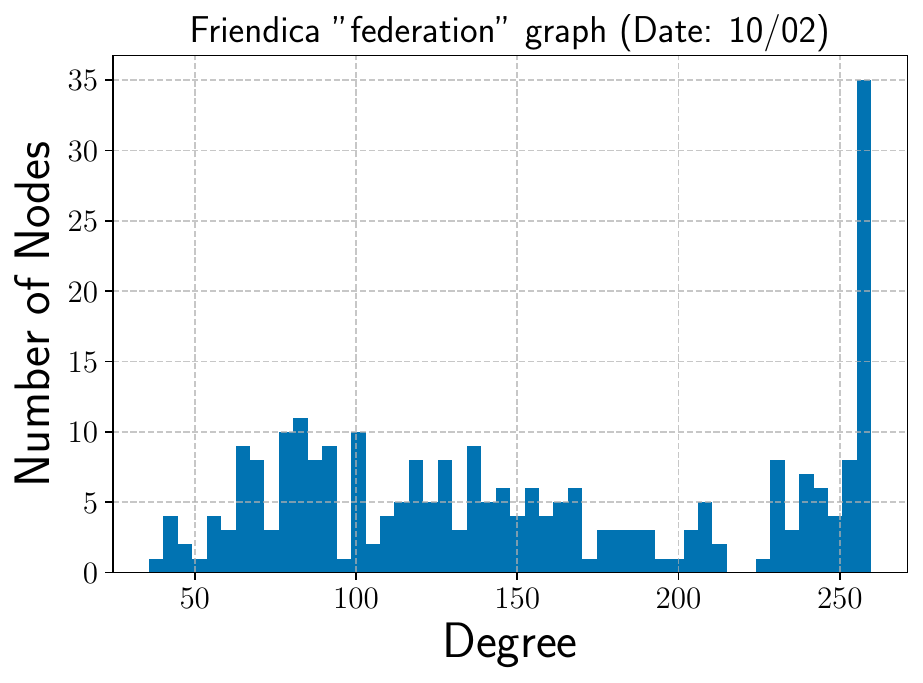}
    \caption{Friendica}
    \label{fig:hist_friendica_fed}
  \end{subfigure}\hfill
  \begin{subfigure}[b]{0.24\textwidth}
    \centering
    \includegraphics[width=\linewidth]{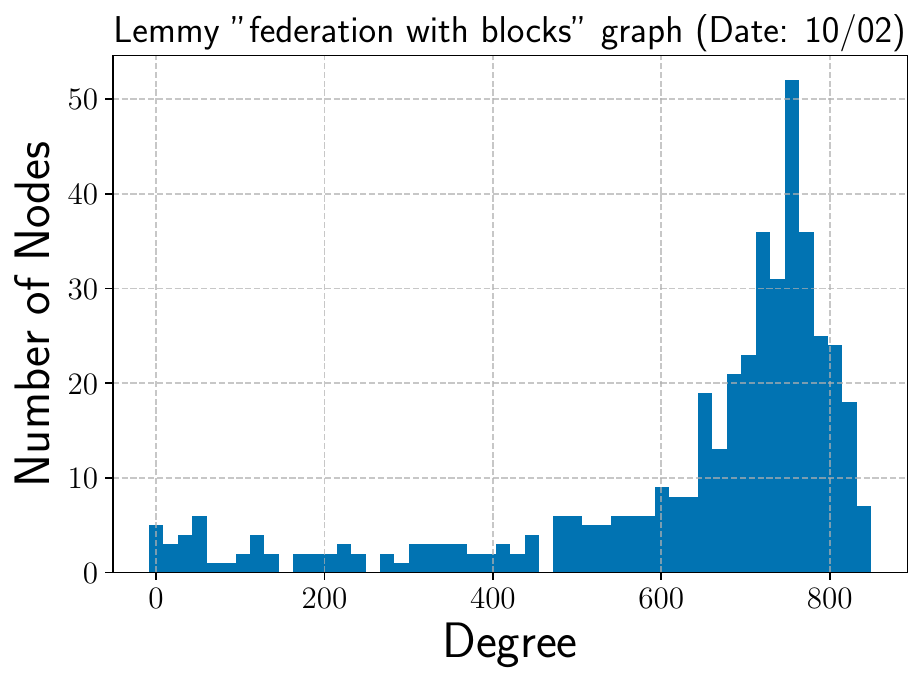}
    \caption{Lemmy (with blocks)}
    \label{fig:hist_lemmy_blocks_fed}
  \end{subfigure}\hfill
  \begin{subfigure}[b]{0.24\textwidth}
    \centering
    \includegraphics[width=\linewidth]{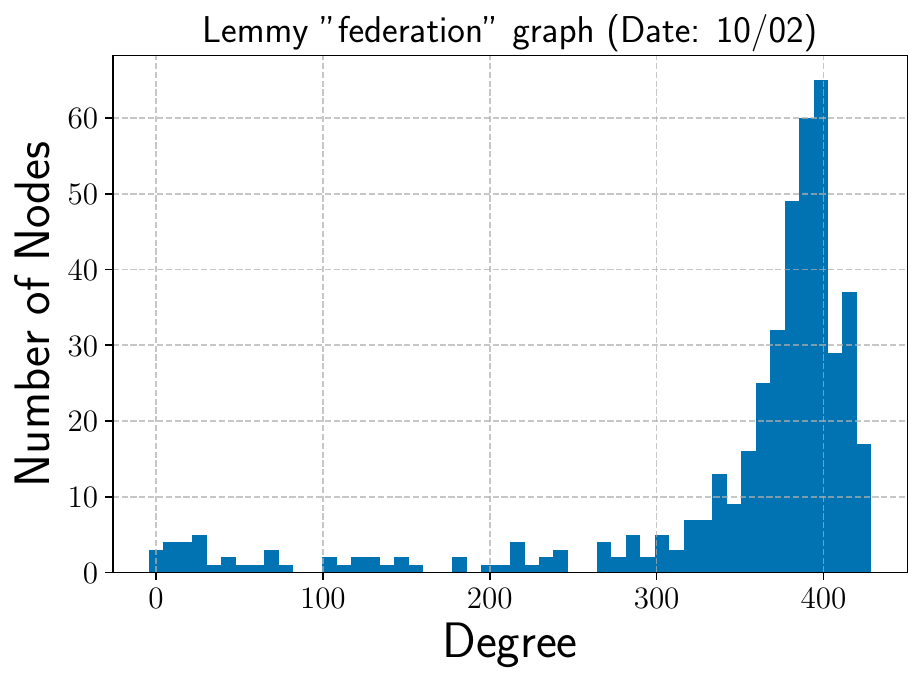}
    \caption{Lemmy}
    \label{fig:hist_lemmy_fed}
  \end{subfigure}

  \vspace{1ex}

  \begin{subfigure}[b]{0.24\textwidth}
    \centering
    \includegraphics[width=\linewidth]{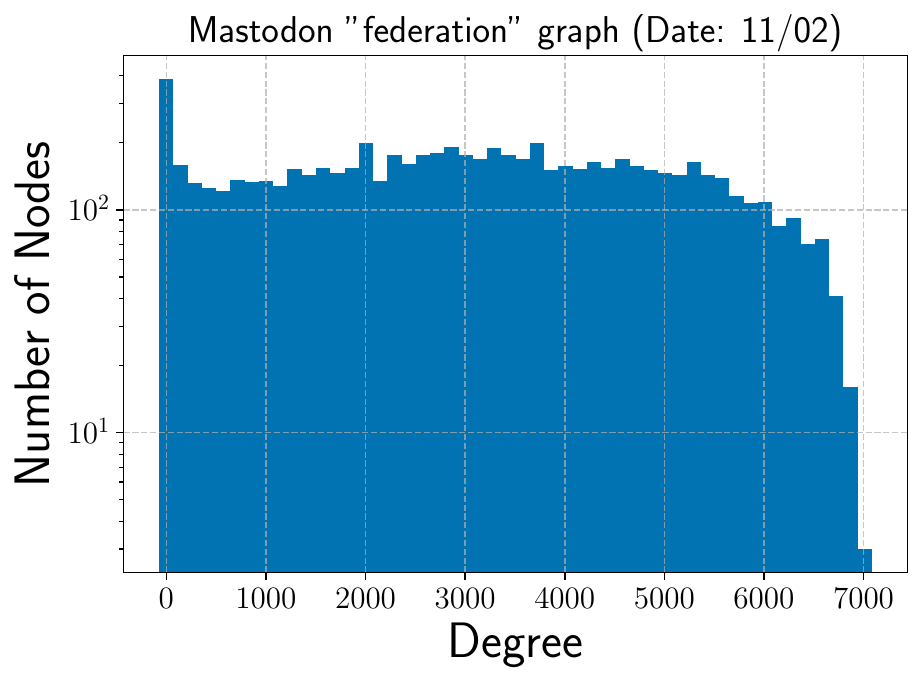}
    \caption{Mastodon}
    \label{fig:hist_mastodon_fed}
  \end{subfigure}\hfill
  \begin{subfigure}[b]{0.24\textwidth}
    \centering
    \includegraphics[width=\linewidth]{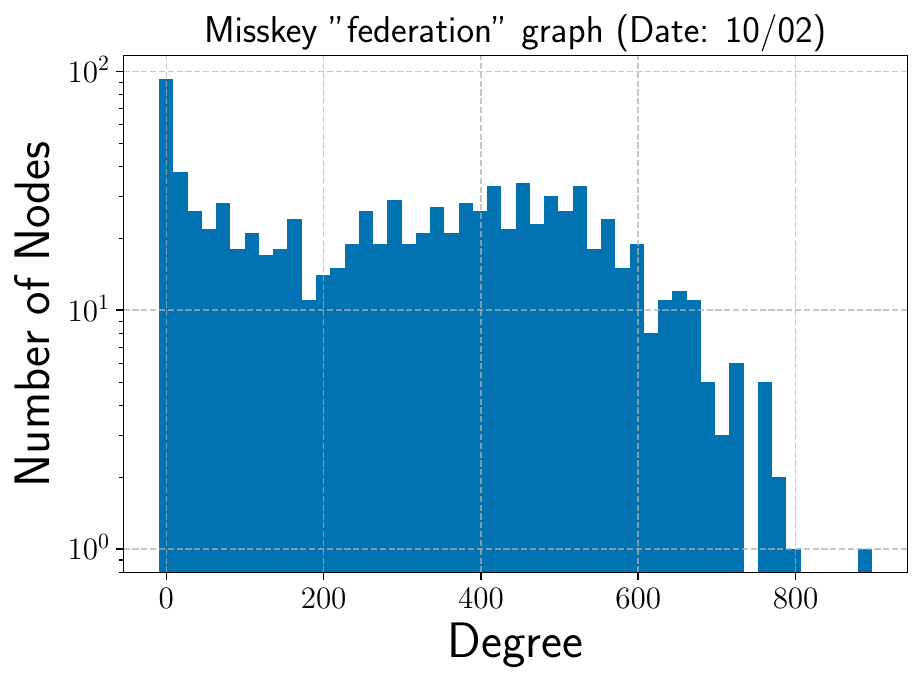}
    \caption{Misskey}
    \label{fig:hist_misskey_fed}
  \end{subfigure}\hfill
  \begin{subfigure}[b]{0.24\textwidth}
    \centering
    \includegraphics[width=\linewidth]{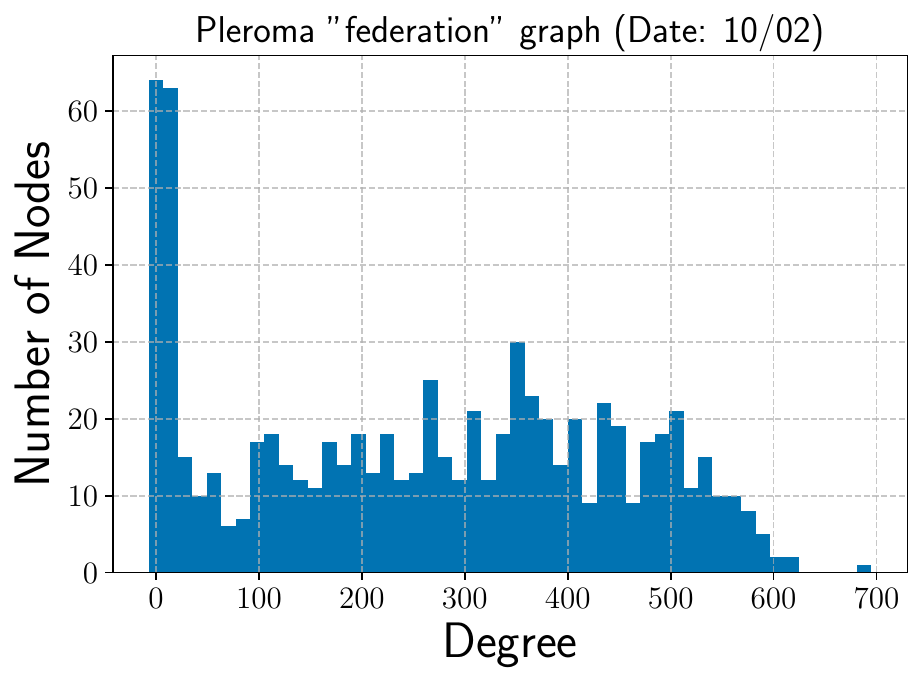}
    \caption{Pleroma}
    \label{fig:hist_pleroma_fed}
  \end{subfigure}\hfill
  \begin{subfigure}[b]{0.24\textwidth}
    \centering
    \includegraphics[width=\linewidth]{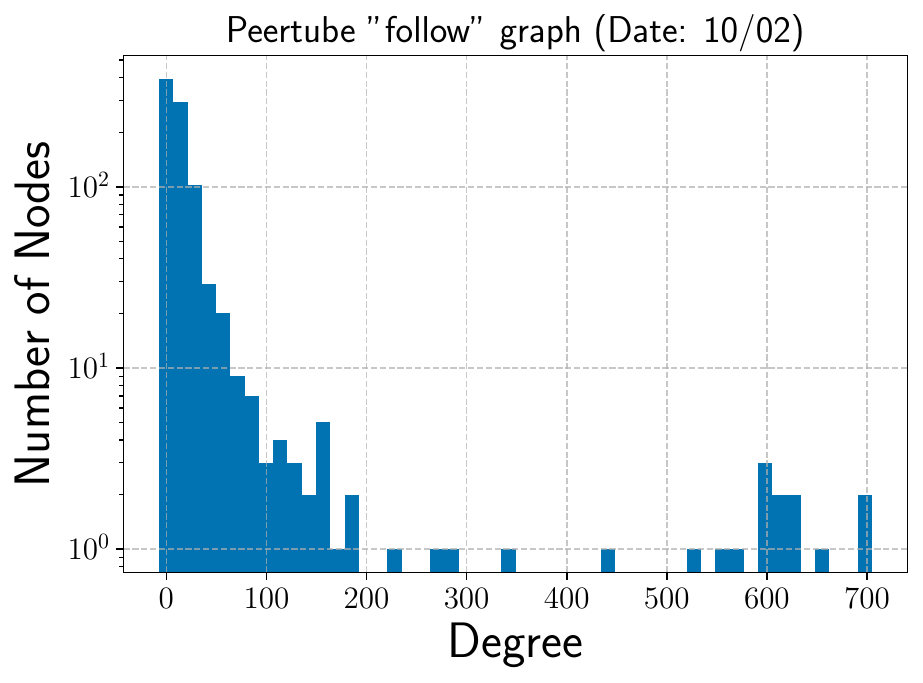}
    \caption{PeerTube}
    \label{fig:hist_peertube_fed}
  \end{subfigure}

  \caption{Degree‐histogram distributions for federation graphs. Version with and without block for Lemmy, and Follow graph for Peertube}
  \label{fig:all_histograms_federation}
\end{figure}

\begin{figure}[htb]
  \centering
  \begin{subfigure}[b]{0.19\textwidth}
    \centering
    \includegraphics[width=\linewidth]{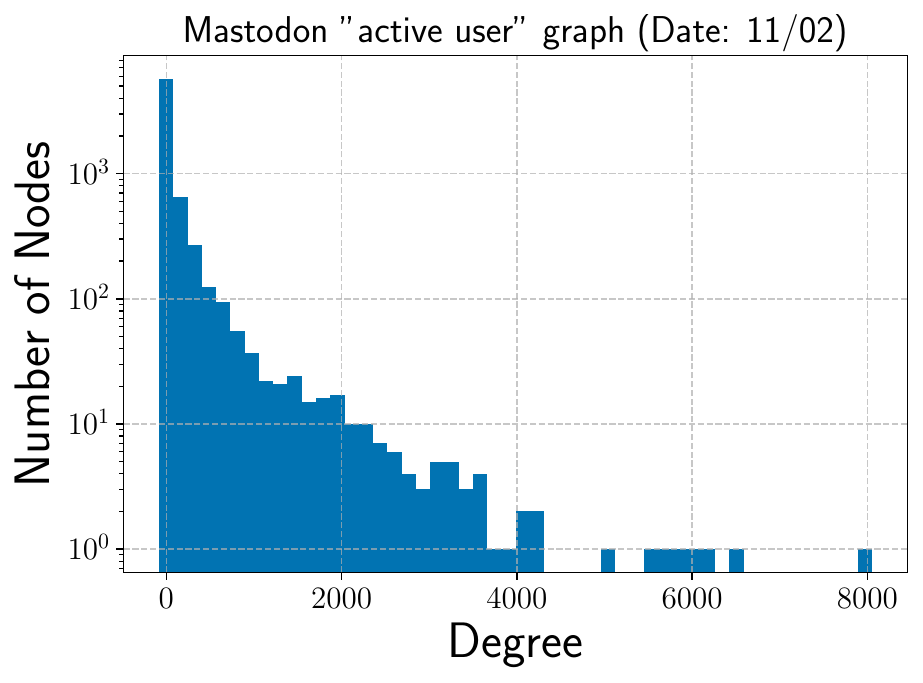}
    \caption{Mastodon\\ (active users)}
    \label{fig:hist_mastodon_active}
  \end{subfigure}\hfill
  \begin{subfigure}[b]{0.19\textwidth}
    \centering
    \includegraphics[width=\linewidth]{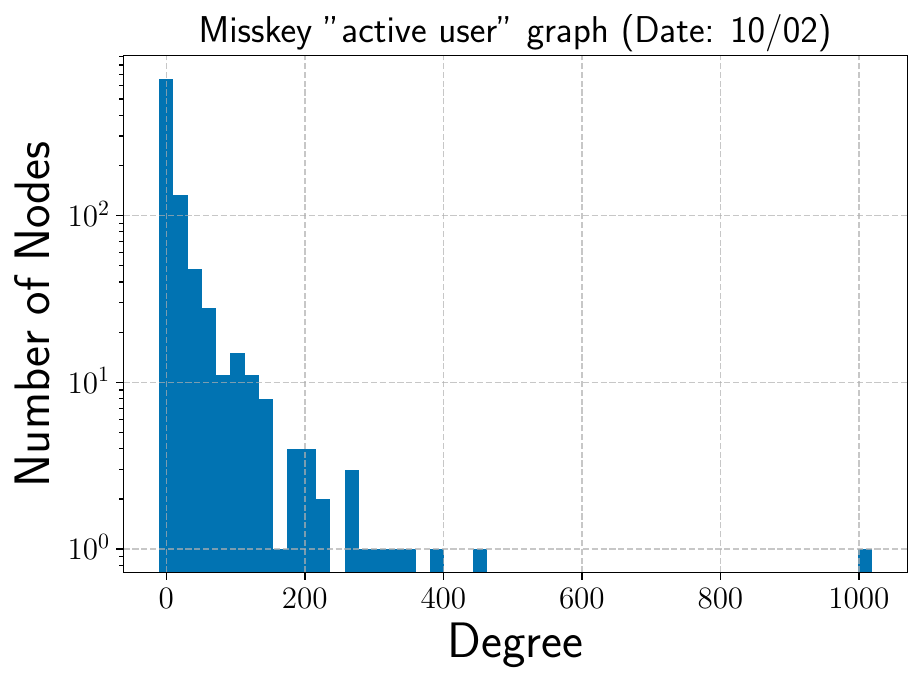}
    \caption{Misskey\\ (active users)}
    \label{fig:hist_misskey_active}
  \end{subfigure}\hfill
  \begin{subfigure}[b]{0.19\textwidth}
    \centering
    \includegraphics[width=\linewidth]{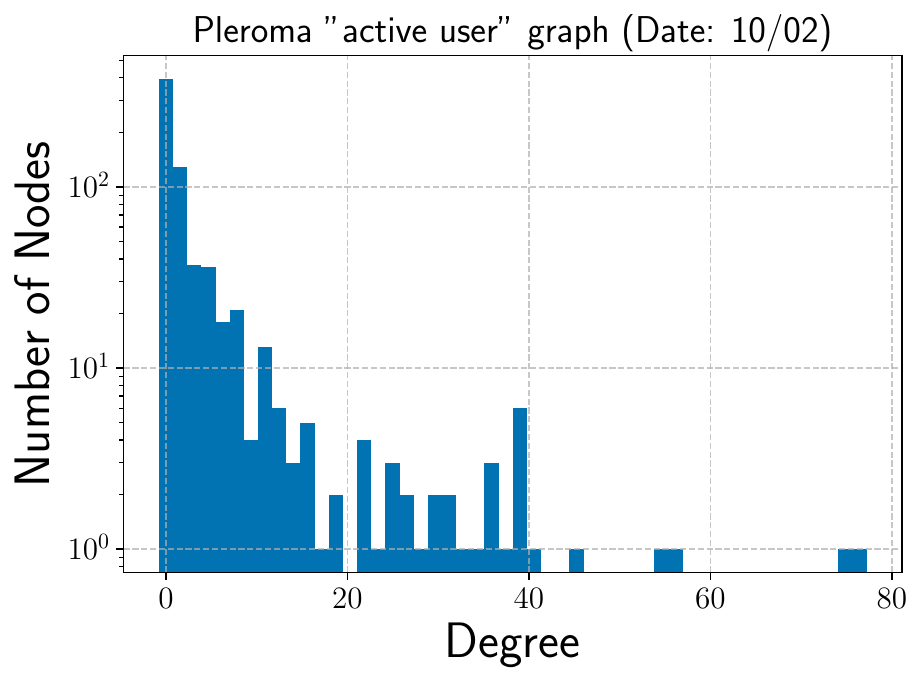}
    \caption{Pleroma\\ (active users)}
    \label{fig:hist_pleroma_active}
  \end{subfigure}\hfill
  \begin{subfigure}[b]{0.19\textwidth}
    \centering
    \includegraphics[width=\linewidth]{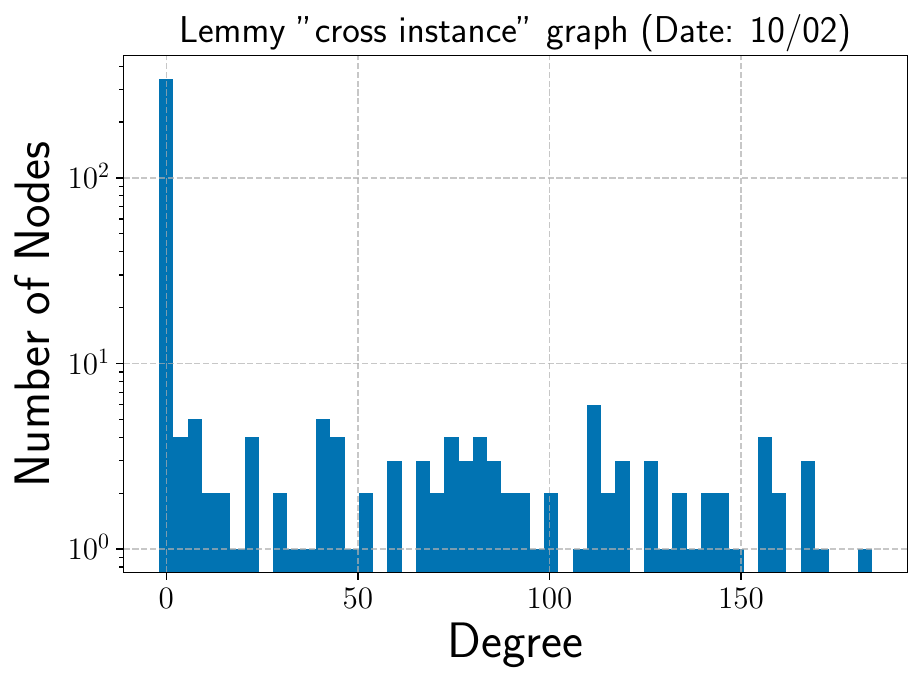}
    \caption{Lemmy\\ (cross-instance)}
    \label{fig:hist_lemmy_cross}
  \end{subfigure}\hfill
  \begin{subfigure}[b]{0.19\textwidth}
    \centering
    \includegraphics[width=\linewidth]{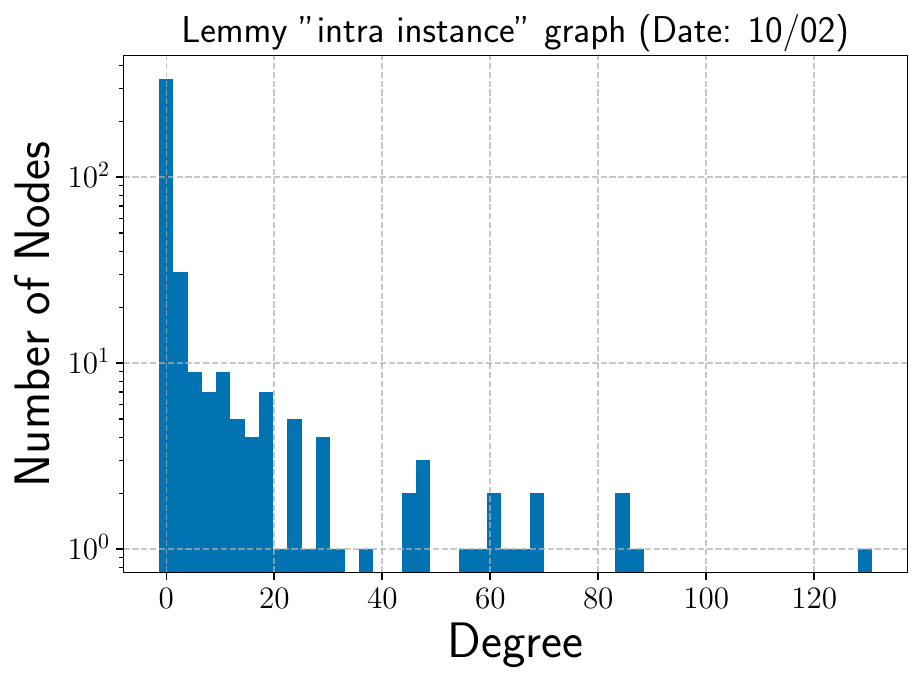}
    \caption{Lemmy\\ (intra-instance)}
    \label{fig:hist_lemmy_intra}
  \end{subfigure}

  \caption{Degree‐histogram distributions for the three active‐user graphs and Lemmy’s cross- vs. intra-instance graphs.}
  \label{fig:hist_remaining}
\end{figure}

\paragraph{Degree distribution}
Fig. \ref{fig:all_histograms_federation} and \ref{fig:hist_remaining} details the degree distribution for each Fediverse software.

For federation graphs, we identify two distinct patterns.  
First, in BookWyrm, Lemmy, and Friendica, the degree distribution exhibits a peak at high degrees, and the proportion of nodes decreases with the degree. This behavior is particularly pronounced in Lemmy.  
In contrast, Mastodon, Misskey, and Pleroma exhibit a peak at low degrees (close to $0$), followed by a relatively flat region for intermediate degrees, and a sharp drop for high degrees.  
This similarity can be attributed to the fact that these platforms are all centered on the same activity: micro-blogging.

In the active user graphs, the degree distribution consistently follows a power-law-like pattern, with the number of nodes decreasing as the degree increases.  
The intra-instance graph of Lemmy exhibits a similar trend, though the pattern is less pronounced.

\begin{figure}
    \centering
    \includegraphics[width=0.5\linewidth]{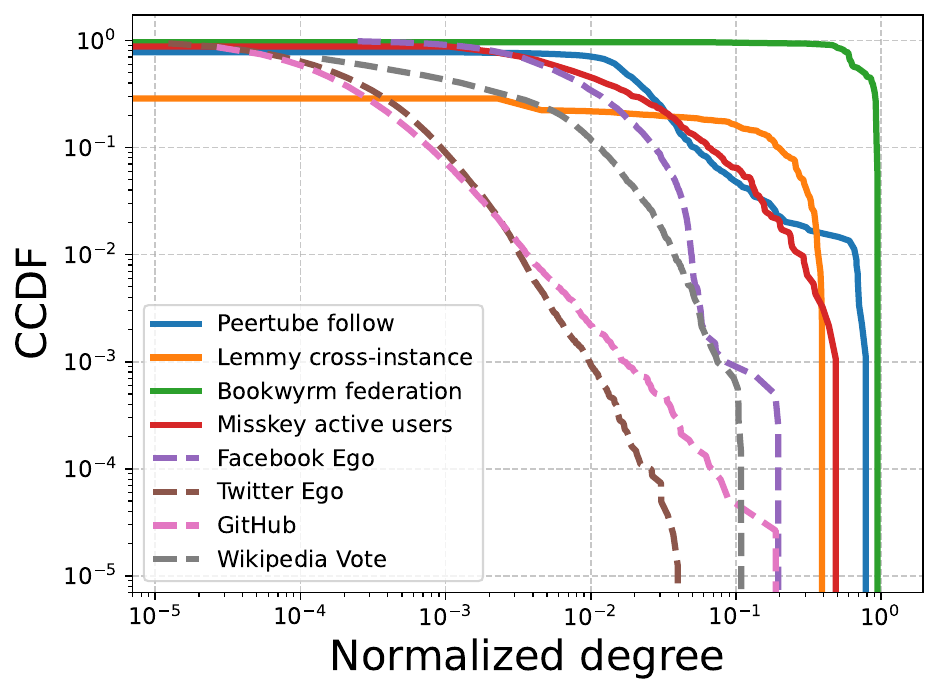}
    \caption{Complementary Cumulative Distribution
Function (CCDF) of the degree for several Fedivertex graphs and other widely-used graphs \cite{leskovec_snap_2014} based on social networks after normalization by the total number of nodes}
    \label{fig:normalizedccdf}
\end{figure}

Fig. \ref{fig:normalizedccdf} reproduces \cref{fig:comparison} from the main text, but with a normalized degree: the degree is divided by the number of nodes.
This normalization incorporates graph density: the closer a curve is to the top-right corner, the denser the graph.
From \cref{fig:normalizedccdf}, Fedivertex graphs appear denser than the SNAP social graphs.  
While the two groups overlap in \cref{fig:comparison}, they are more distinctly separated in the normalized plot.  
This further illustrates major differences between Fedivertex and widely-used datasets.

\begin{figure}[htb]
  \centering
  \begin{subfigure}[b]{0.42\textwidth}
    \centering
    \includegraphics[width=\linewidth]{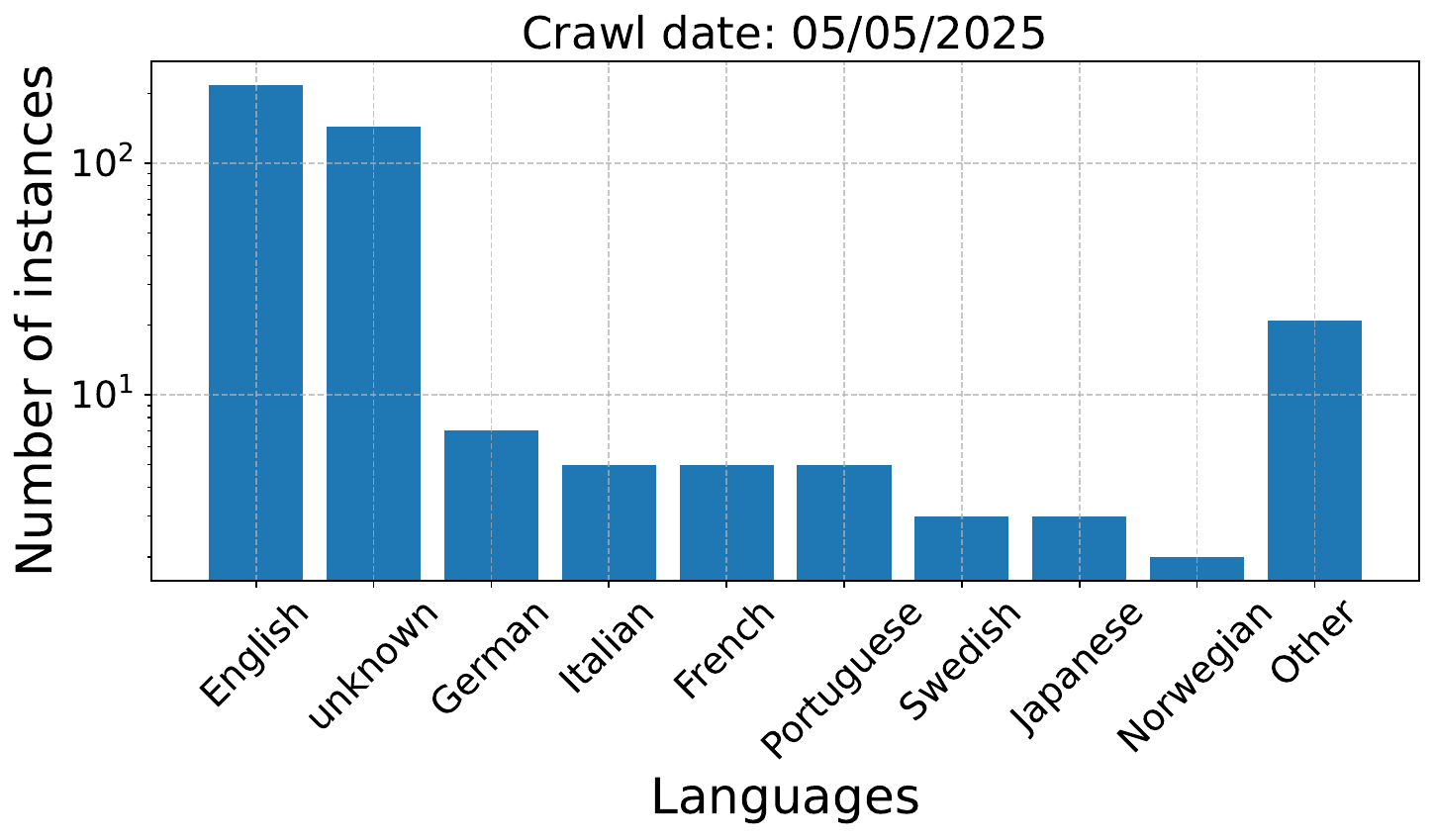}
    \caption{Lemmy}
    \label{fig:lang_hist_lemmy}
  \end{subfigure}\hspace{2em}
  \begin{subfigure}[b]{0.42\textwidth}
    \centering
    \includegraphics[width=\linewidth]{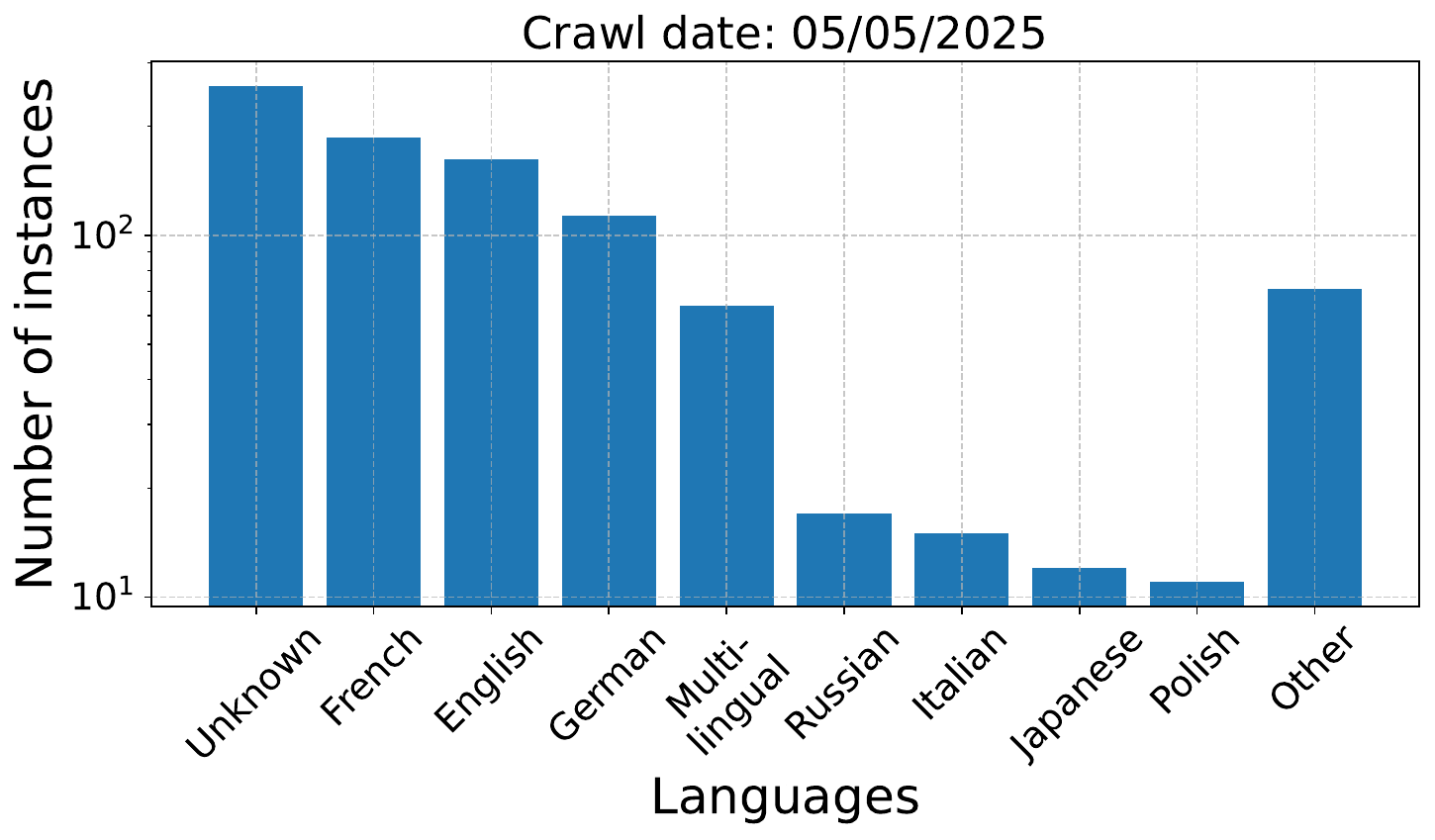}
    \caption{Peertube}
    \label{fig:lang_hist_peertube}
  \end{subfigure}\\\vspace{1em}
  \begin{subfigure}[b]{0.42\textwidth}
    \centering
    \includegraphics[width=\linewidth]{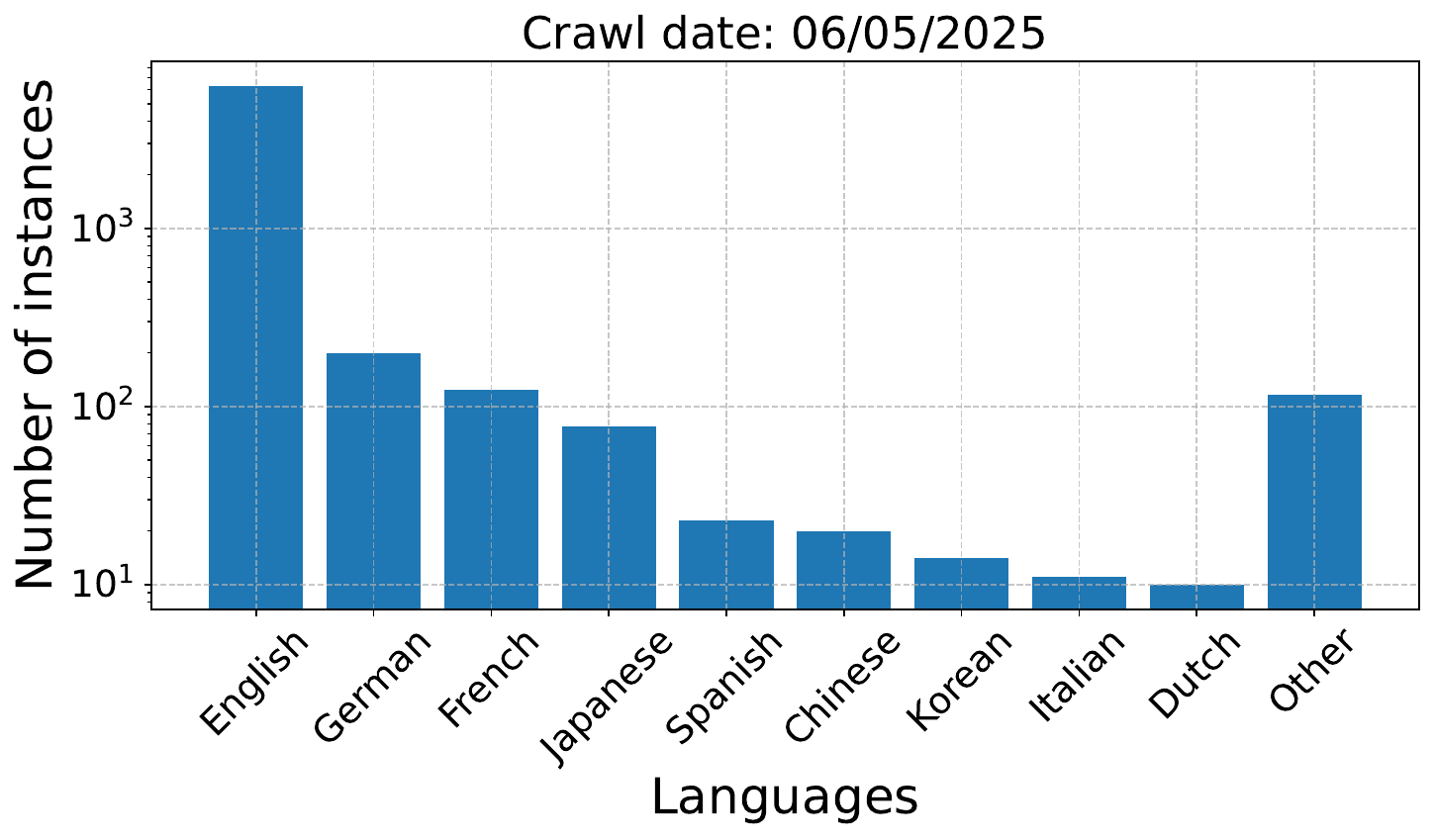}
    \caption{Mastodon}
    \label{fig:lang_hist_mastodon}
  \end{subfigure}\hspace{2em}
  \begin{subfigure}[b]{0.42\textwidth}
    \centering
    \includegraphics[width=\linewidth]{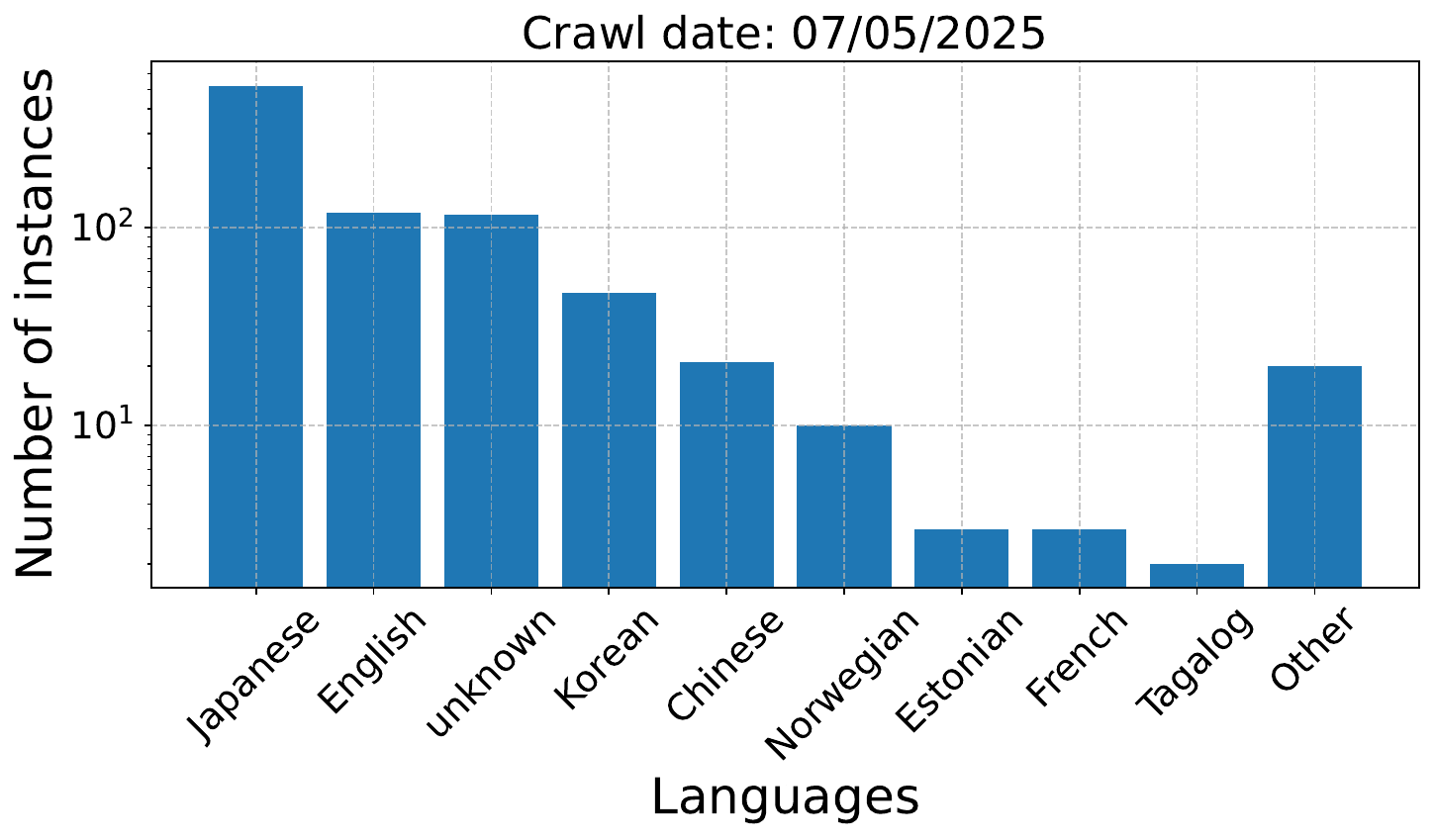}
    \caption{Misskey}
    \label{fig:lang_hist_misskey}
  \end{subfigure}
  \caption{Histograms of the language distribution among four different Fediverse software, logarithmic scale}
  \label{fig:lang_hist}
\end{figure}

\paragraph{Language distribution}
Fig. \ref{fig:lang_hist} presents histograms of the language distribution in Lemmy, Peertube, Mastodon, and Misskey.
This figure shows diverse language distributions among Fediverse social media.
First, Mastodon and Lemmy have a large majority of English-speaking instances.
Second, Peertube is dominated by European languages.
We observe more diversity than in Mastodon because French and German have a number of nodes similar to English.
Third, Misskey shows a unique language distribution within the Fediverse as the Asian languages (especially Japanese) are the most spoken languages.
This phenomenon is explained by the fact that Misskey (contrary to other Fediverse software) has been developed by a Japanese developer.

\paragraph{For a thorough exploratory data analysis} we refer to our 5 Kaggle notebooks that cover these aspects, but also other dimensions such as weight or user distribution:\\\url{https://www.kaggle.com/datasets/marcdamie/fediverse-graph-dataset/code}.

\section{Python API examples}
\label{app:code}
Our Python package \texttt{Fedivertex} produces a straightforward interface to interact with our dataset.
This package seamlessly downloads and extracts graphs from our CSV files using the NetworkX format, a popular Python package to manipulate graphs.
This appendix demonstrates several simple operations that can be performed using our package.

The installation is straightforward as our package is available on the Python Package Index (PyPI): \texttt{pip3 install fedivertex}.
The dataset is downloaded automatically the first time a graph is loaded.
The dataset is stored in cache thanks to the \texttt{CroissantML} package in the folder \texttt{.cache/croissant}.

\begin{listing}[tb]
\tiny
\begin{minted}[frame=single,]{Python}
from fedivertex import GraphLoader

loader = GraphLoader()
G = loader.get_graph(software = "misskey", graph_type = "active_user")

nb_nodes = G.number_of_nodes()
nb_edges = G.number_of_edges()
degree = G.degree()
\end{minted}
\caption{Code example to extract statistics about a Fedivertex graph}
\label{lst:basic_statistics}
\end{listing}

Listing \ref{lst:basic_statistics} shows how to extract basic statistics from the latest Misskey active user graph.
Our method \texttt{get\_graph} outputs a NetworkX object, so we can apply all functions and methods from NetworkX.
This example simply calls \texttt{number\_of\_nodes}, \texttt{number\_of\_edges}, and \texttt{degree}, but one can use the entire NetworkX API.
We refer to NetworkX documentation for further information about the capabilities of their API: \url{https://networkx.org/documentation/stable/index.html}

\begin{listing}[tb]
\tiny
\begin{minted}[frame=single,]{Python}
from fedivertex import GraphLoader
loader = GraphLoader()

# Extract the latest graph
G_latest = loader.get_graph(software = "misskey", graph_type = "active_user")

# Extract the i-th graph in the dataset (sorted chronologically)
i = 3
G_i = loader.get_graph(software = "misskey", graph_type = "active_user", index=i)

# Extract the graph of a specific data
G_2403 = loader.get_graph(software = "misskey", graph_type = "active_user", date="20250324")
\end{minted}
\caption{Code example of the graph selection in Fedivertex}
\label{lst:graph_selection}
\end{listing}

In Fedivertex, a graph is defined using three elements: the software (e.g., Misskey), the graph type (e.g., active user), and the date.
For date selection, the package provides three different options, illustrated in \cref{lst:graph_selection}.
First, if no date is provided to \texttt{get\_graph}, the latest graph is automatically selected.
Second, if the user provides an (integer) index $i$, the $i$-th graph is selected.
This index works like a Python list index, so $-1$ is accepted and the indexing starts at $0$.
Third, the user can provide a date in a string format \texttt{YYYYMMDD} (e.g., $20250324$).

\begin{listing}[tb]
\tiny
\begin{minted}[frame=single,]{Python}
from fedivertex import GraphLoader
loader = GraphLoader()

print(loader.list_all_software())
# ["bookwyrm", "friendica", "lemmy", "mastodon", "misskey", "peertube", "pleroma"]

print(loader.list_graph_types("peertube"))
# ["follow"]

print(loader.list_available_dates("peertube", "follow"))
# ["20250203", "20250210", "20250217", "20250224", "20250303", "20250311",    "20250317",
#    "20250324", "20250331", "20250407", "20250414", "20250421", "20250428", "20250505"]
\end{minted}
\caption{Code example of the utility functions in Fedivertex}
\label{lst:utility_functions}
\end{listing}

As our dataset contains a large range of possible software, graph types, and dates, we also provide utility functions to simplify the interactions.
Listing \ref{lst:utility_functions} presents these functions.
First, \texttt{list\_all\_software} lists all the available software.
Second, \texttt{list\_graph\_types} lists the available graph types for a given software.
Third, \texttt{list\_available\_dates} lists the available dates for a given software and graph type.

As seen in \cref{list:code_example} in main text, it is possible to use graph extraction with the option ``\texttt{only\_largest\_component = True}.''
When set to true, the method only returns the largest component of the graph.
As many graph algorithms require connected graphs, this option eases benchmarks of such algorithms.
Additionally, \cref{lst:language_extraction} shows how to extract the language information to use it as a ground-truth label (e.g., in community detection like in \cref{tab:community-detection})

For further advanced examples, we refer to our Kaggle notebooks that analyze the dataset using the Python package:\\\url{https://www.kaggle.com/datasets/marcdamie/fediverse-graph-dataset/code}.

\section{Experiments on defederation}
\label{app:defederation}

\begin{figure}[htb]
  \centering
  \begin{subfigure}[b]{0.24\textwidth}
    \centering
    \includegraphics[width=\linewidth]{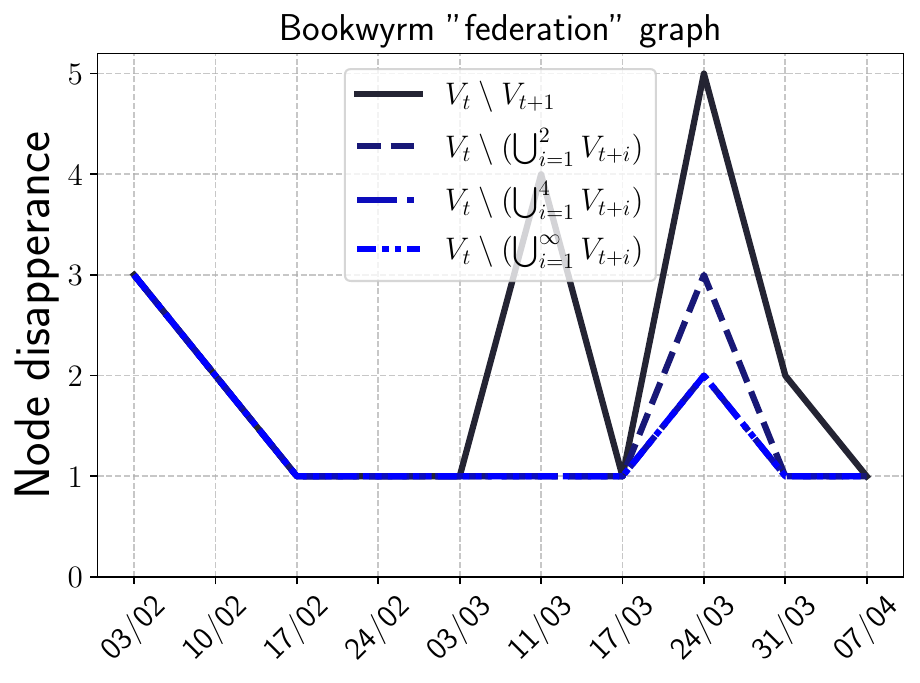}
    \caption{BookWyrm}
    \label{fig:node_deletion_bookwyrm}
  \end{subfigure}\hfill
  \begin{subfigure}[b]{0.24\textwidth}
    \centering
    \includegraphics[width=\linewidth]{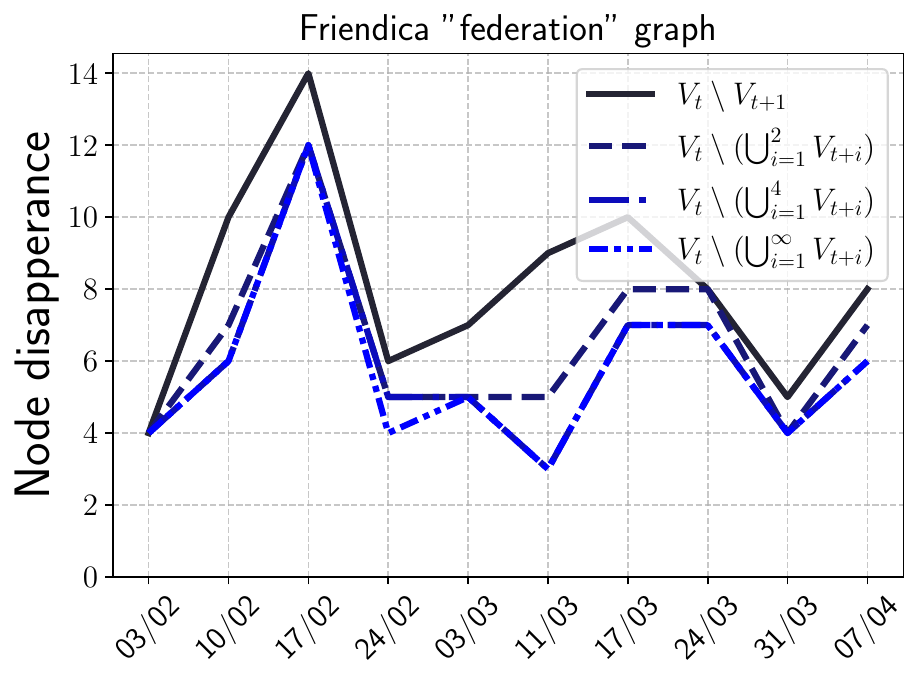}
    \caption{Friendica}
    \label{fig:node_deletion_friendica}
  \end{subfigure}\hfill
  \begin{subfigure}[b]{0.24\textwidth}
    \centering
    \includegraphics[width=\linewidth]{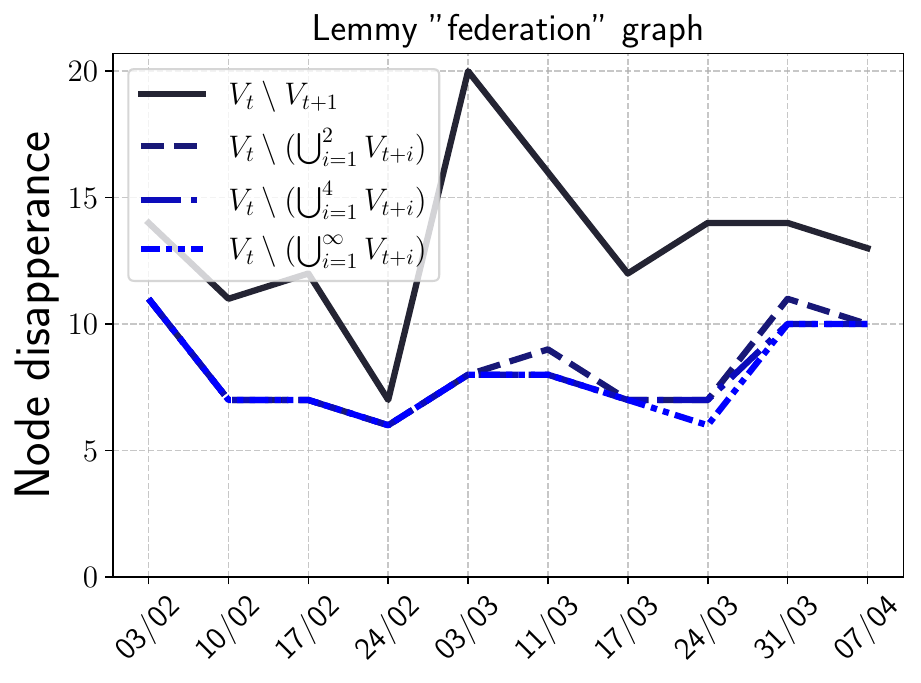}
    \caption{Lemmy}
    \label{fig:node_deletion_lemmy}
  \end{subfigure}\hfill
  \begin{subfigure}[b]{0.24\textwidth}
    \centering
    \includegraphics[width=\linewidth]{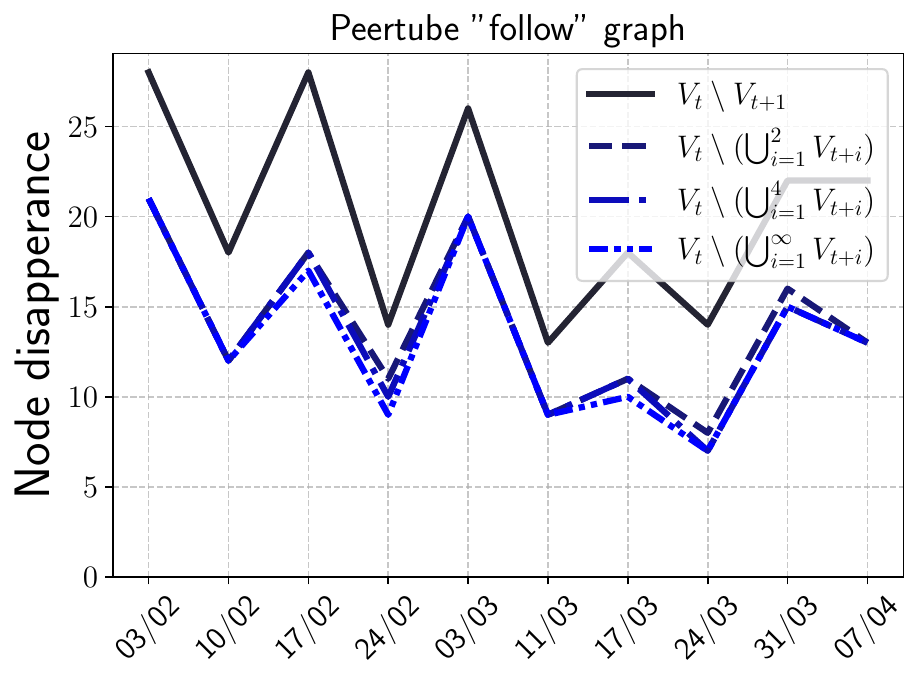}
    \caption{PeerTube}
    \label{fig:node_deletion_peertube}
  \end{subfigure}

  \vspace{1ex}

  \begin{subfigure}[b]{0.24\textwidth}
    \centering
    \includegraphics[width=\linewidth]{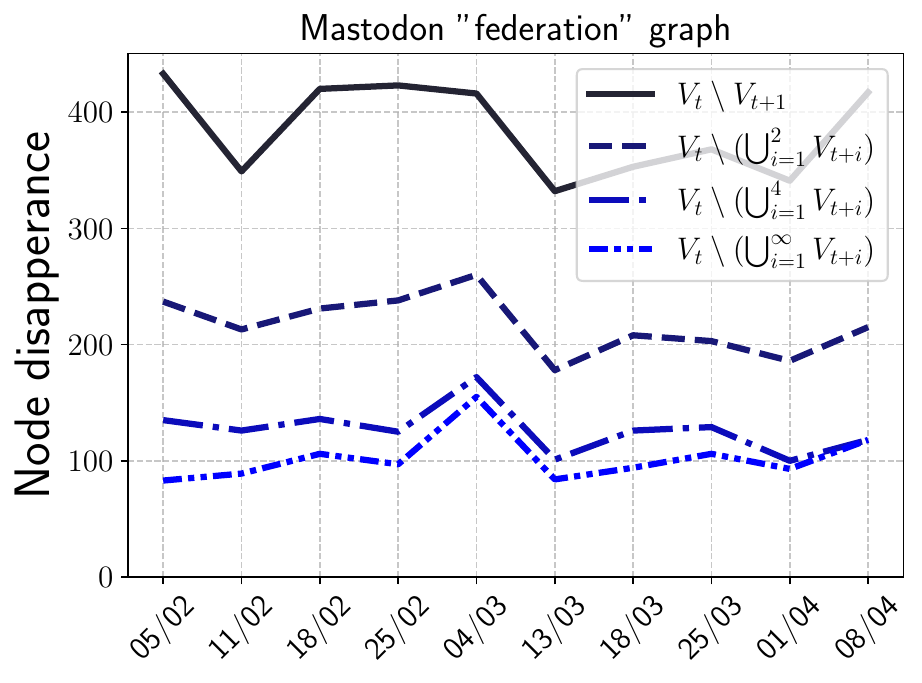}
    \caption{Mastodon}
    \label{fig:node_deletion_mastodon}
  \end{subfigure}\hfill
  \begin{subfigure}[b]{0.24\textwidth}
    \centering
    \includegraphics[width=\linewidth]{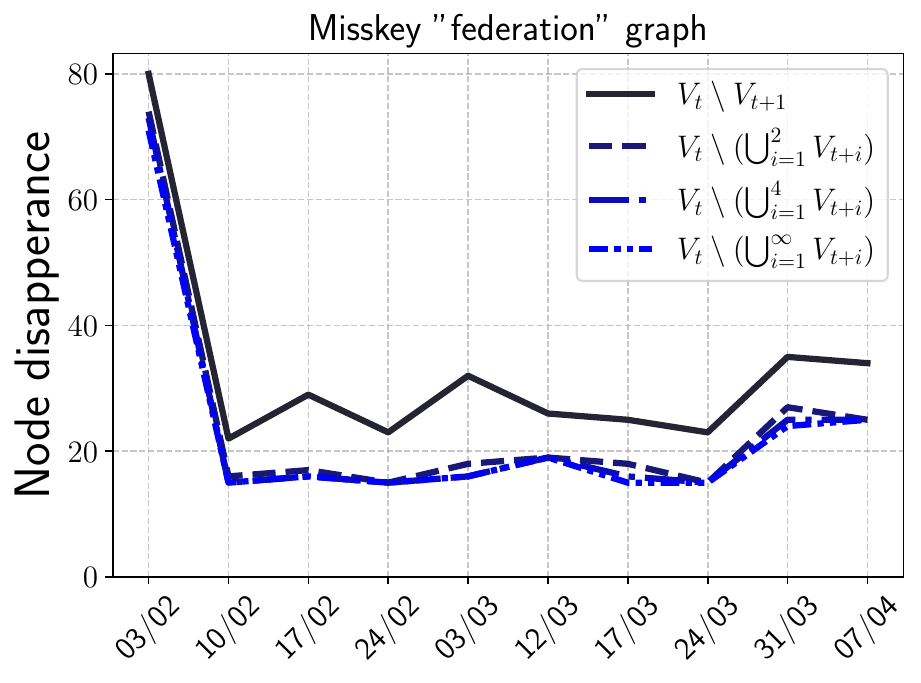}
    \caption{Misskey}
    \label{fig:node_deletion_misskey}
  \end{subfigure}\hfill
  \begin{subfigure}[b]{0.24\textwidth}
    \centering
    \includegraphics[width=\linewidth]{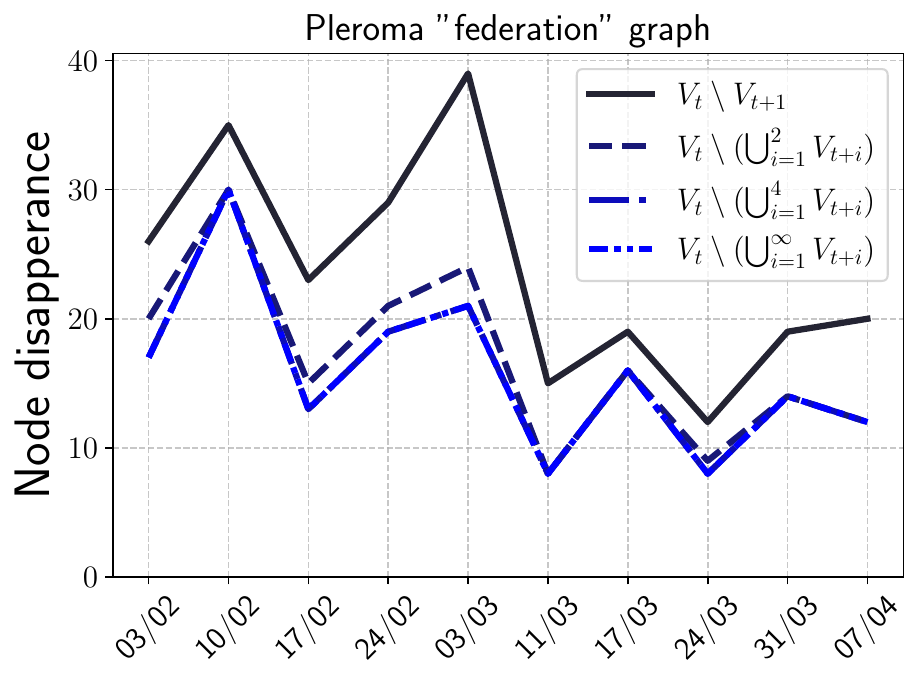}
    \caption{Pleroma}
    \label{fig:node_deletion_pleroma}
  \end{subfigure}
  
  \caption{Node deletion at different time horizons for federation graphs, and the follow graph (for PeerTube).}
  \label{fig:node_deletion}
\end{figure}

\begin{figure}[htb]
  \centering
  \begin{subfigure}[b]{0.24\textwidth}
    \centering
    \includegraphics[width=\linewidth]{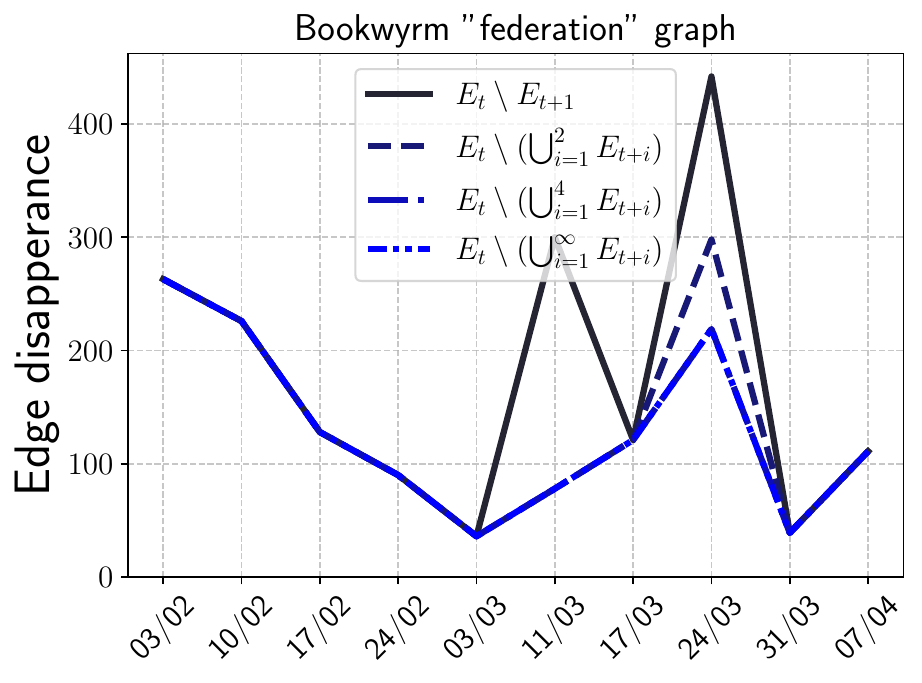}
    \caption{BookWyrm}
    \label{fig:edge_deletion_bookwyrm}
  \end{subfigure}\hfill
  \begin{subfigure}[b]{0.24\textwidth}
    \centering
    \includegraphics[width=\linewidth]{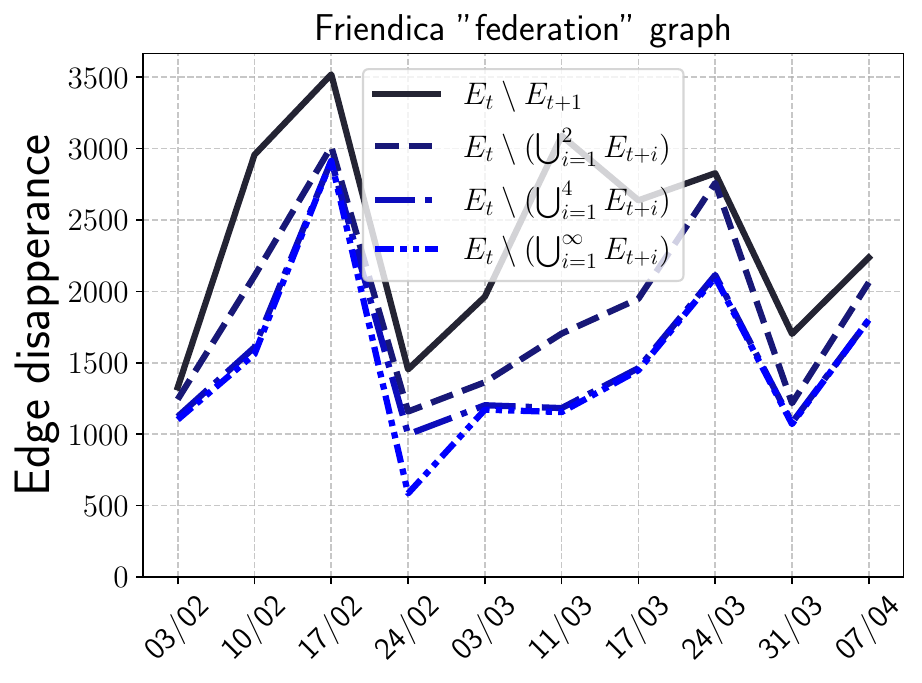}
    \caption{Friendica}
    \label{fig:edge_deletion_friendica}
  \end{subfigure}\hfill
  \begin{subfigure}[b]{0.24\textwidth}
    \centering
    \includegraphics[width=\linewidth]{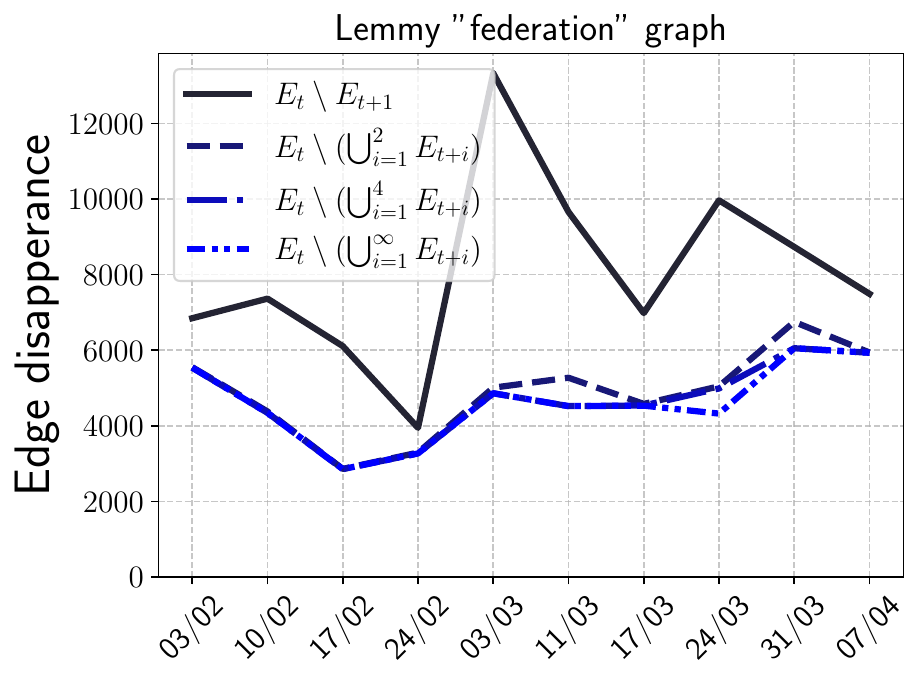}
    \caption{Lemmy}
    \label{fig:edge_deletion_lemmy}
  \end{subfigure}\hfill
   \begin{subfigure}[b]{0.24\textwidth}
    \centering
    \includegraphics[width=\linewidth]{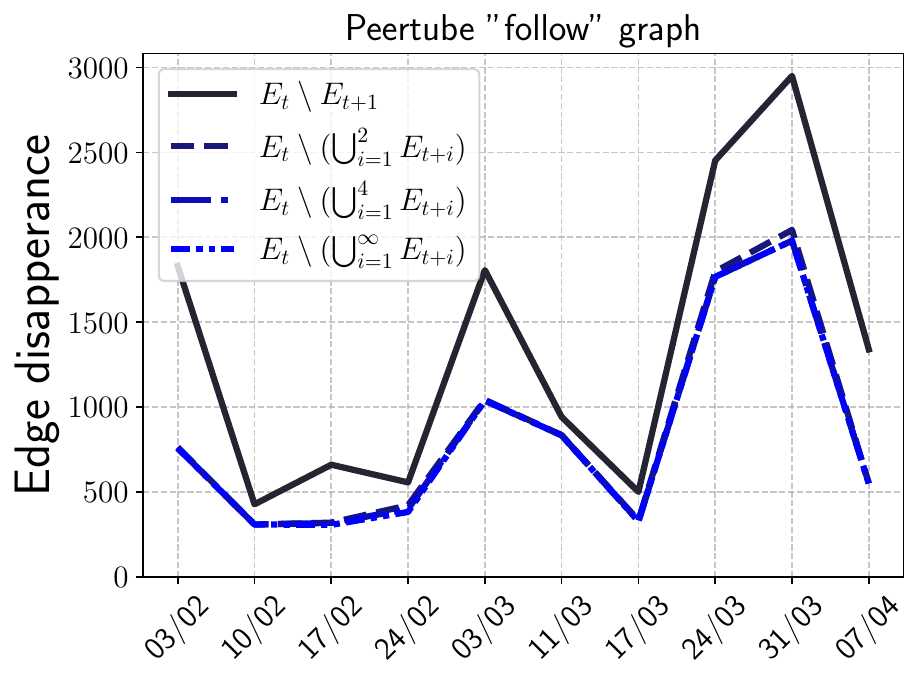}
    \caption{PeerTube}
    \label{fig:edge_deletion_peertube}
  \end{subfigure}
  
  \vspace{1ex}

  \begin{subfigure}[b]{0.24\textwidth}
    \centering
    \includegraphics[width=\linewidth]{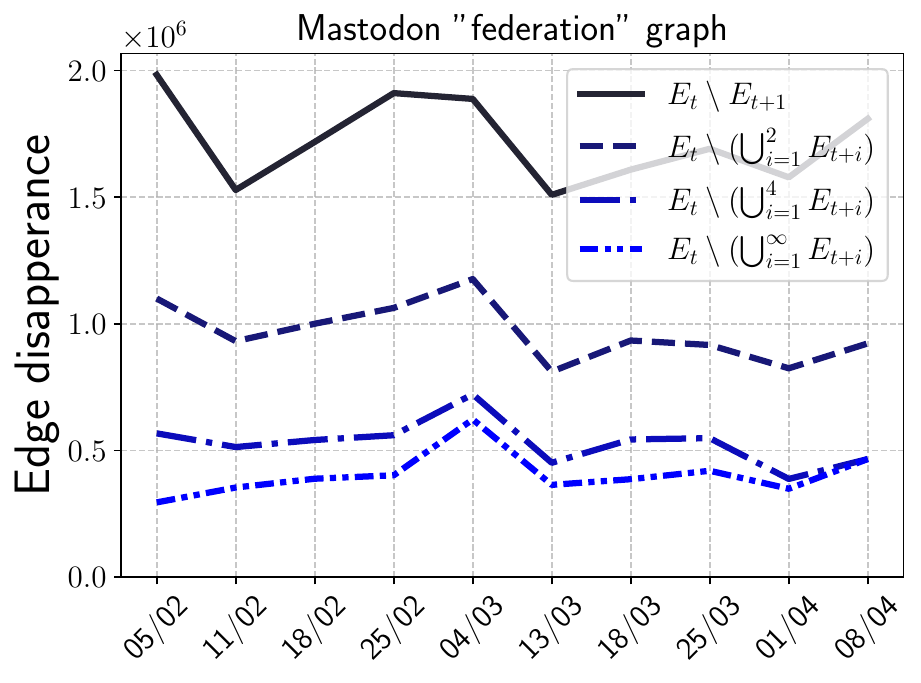}
    \caption{Mastodon}
    \label{fig:edge_deletion_mastodon}
  \end{subfigure}\hfill
  \begin{subfigure}[b]{0.24\textwidth}
    \centering
    \includegraphics[width=\linewidth]{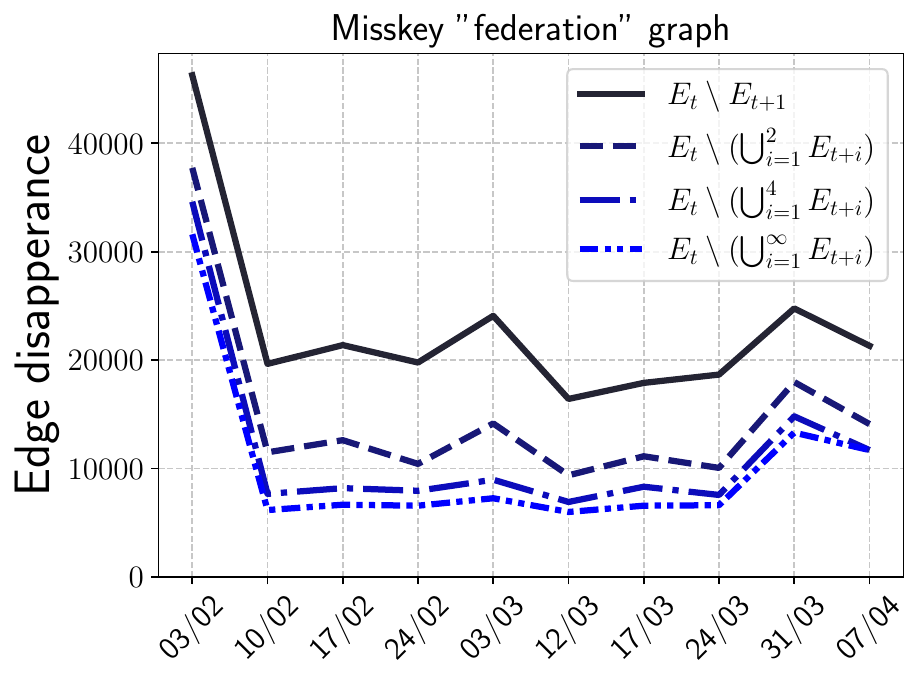}
    \caption{Misskey}
    \label{fig:edge_deletion_misskey}
  \end{subfigure}\hfill
  \begin{subfigure}[b]{0.24\textwidth}
    \centering
    \includegraphics[width=\linewidth]{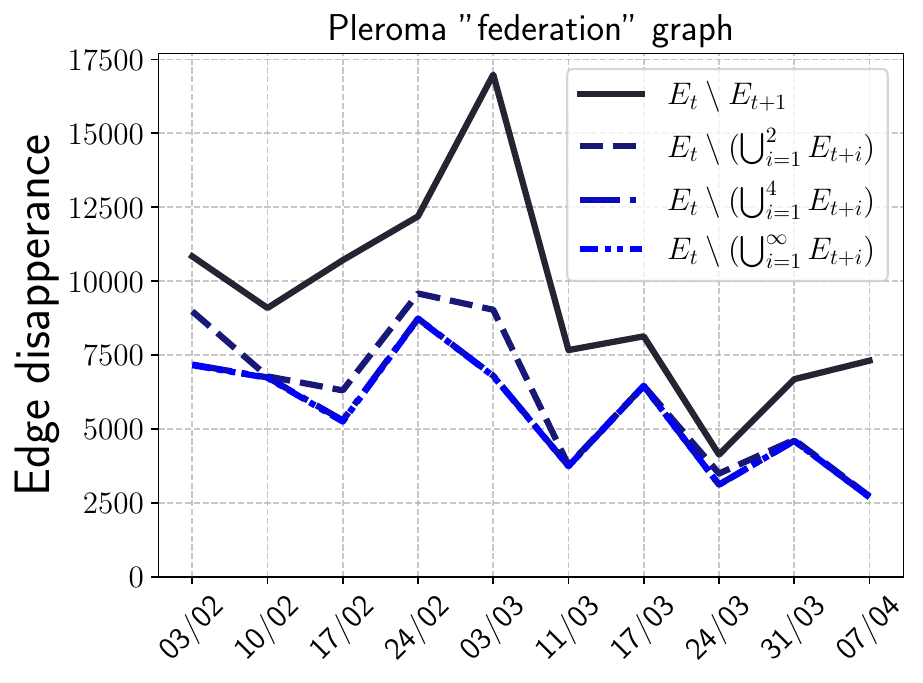}
    \caption{Pleroma}
    \label{fig:edge_deletion_pleroma}
  \end{subfigure}
  
  \caption{Edge deletion at different time horizons for federation graphs, and the follow graph (for PeerTube).}
  \label{fig:edge_deletion}
\end{figure}

\begin{figure}[htb]
  \centering
  \begin{subfigure}[b]{0.19\textwidth}
    \centering
    \includegraphics[width=\linewidth]{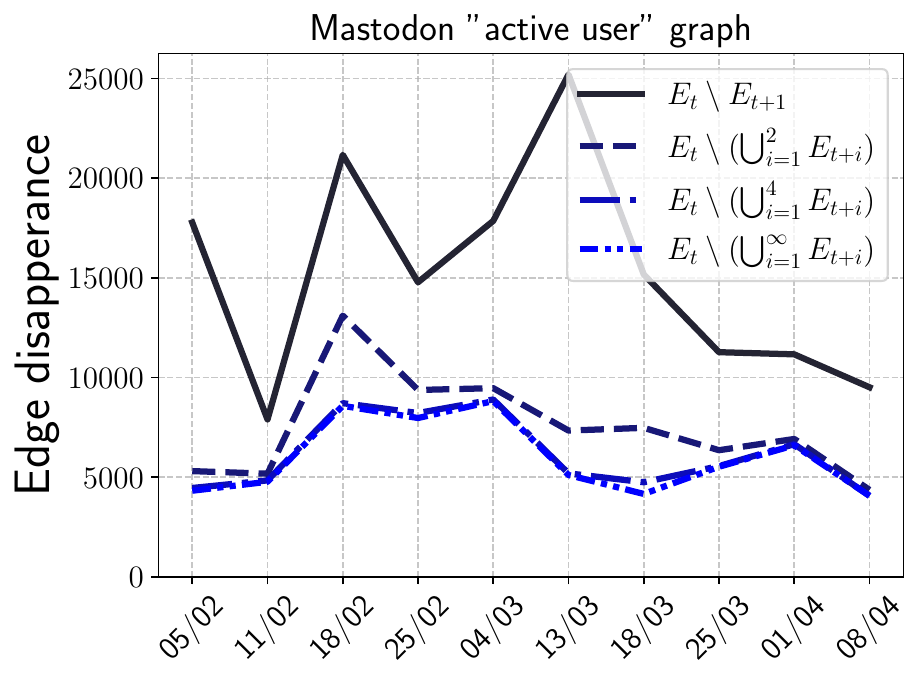}
    \caption{Mastodon\\ (active users)}
    \label{fig:edge_deletion_mastodon_active}
  \end{subfigure}\hfill
  \begin{subfigure}[b]{0.19\textwidth}
    \centering
    \includegraphics[width=\linewidth]{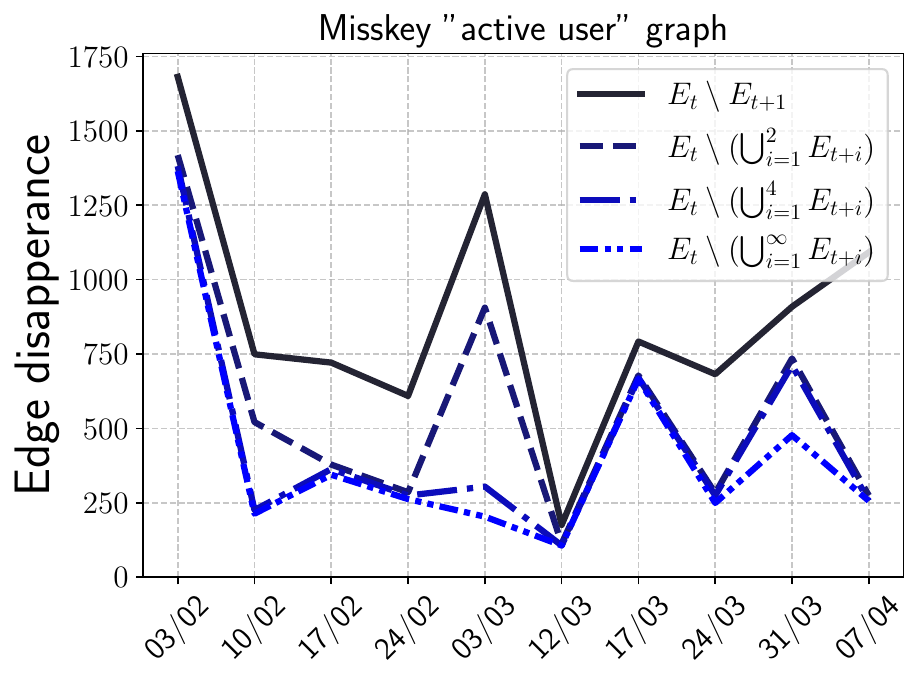}
    \caption{Misskey\\ (active users)}
    \label{fig:edge_deletion_misskey_active}
  \end{subfigure}\hfill
  \begin{subfigure}[b]{0.19\textwidth}
    \centering
    \includegraphics[width=\linewidth]{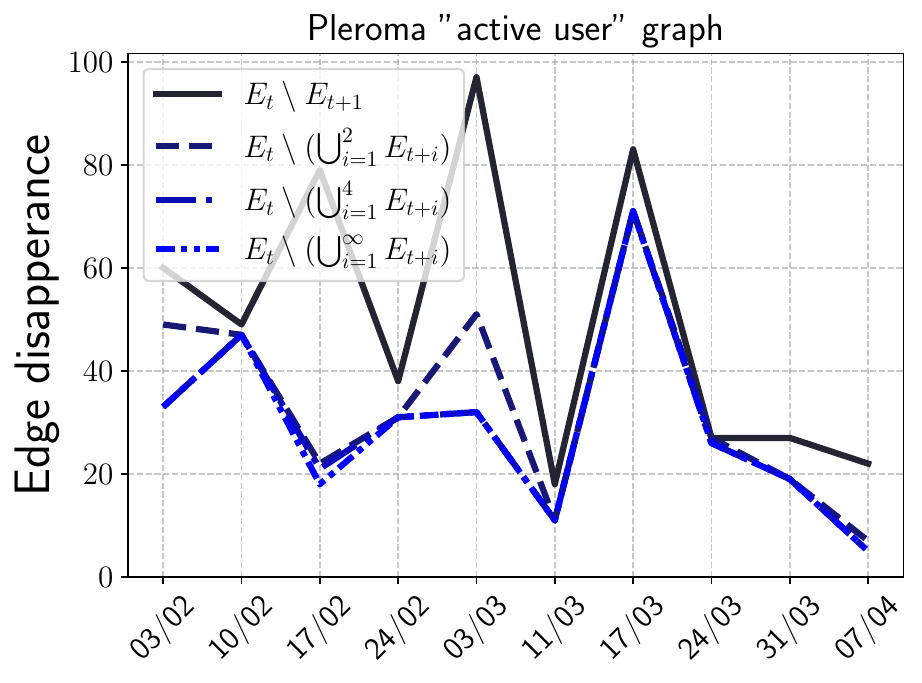}
    \caption{Pleroma\\ (active users)}
    \label{fig:edge_deletion_pleroma_active}
  \end{subfigure}\hfill
  \begin{subfigure}[b]{0.19\textwidth}
    \centering
    \includegraphics[width=\linewidth]{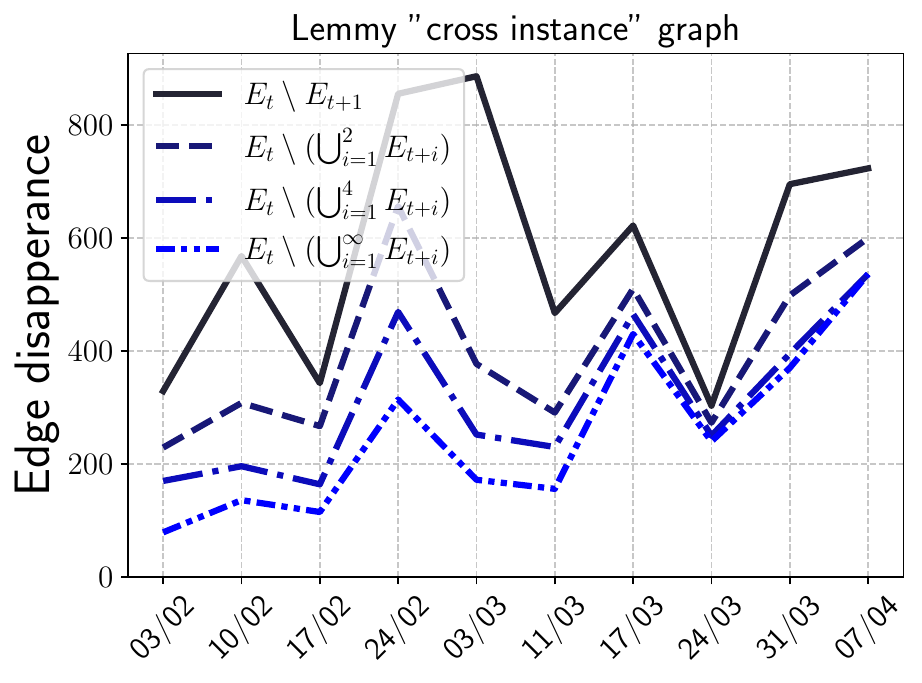}
    \caption{Lemmy\\ (cross-instance)}
    \label{fig:edge_deletion_lemmy_cross}
  \end{subfigure}\hfill
  \begin{subfigure}[b]{0.19\textwidth}
    \centering
    \includegraphics[width=\linewidth]{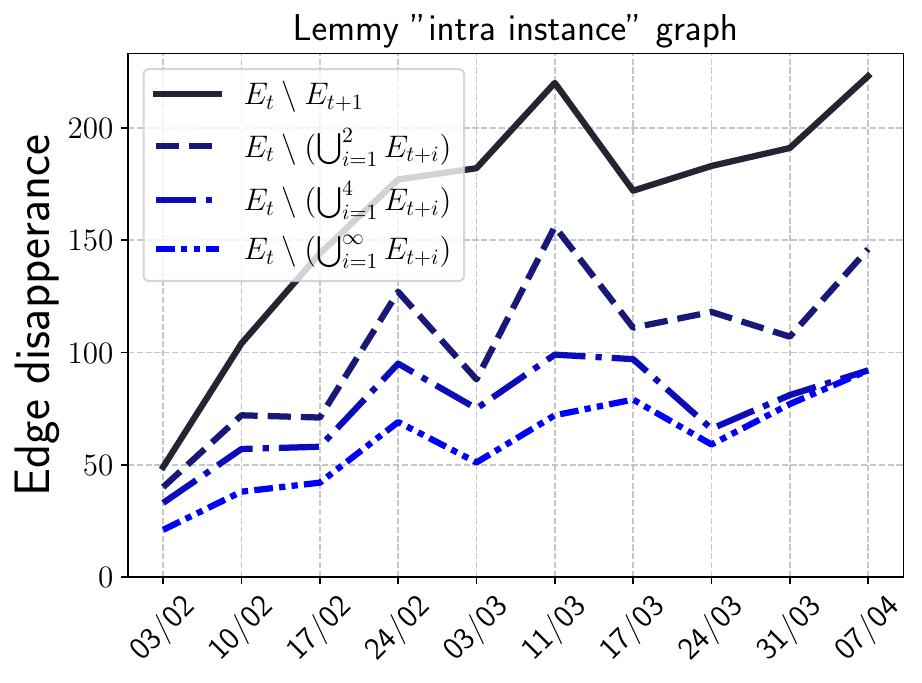}
    \caption{Lemmy\\ (intra-instance)}
    \label{fig:edge_deletion_lemmy_intra}
  \end{subfigure}

  \caption{Edge deletion at different time horizons for the three active‐user graphs and Lemmy’s cross- vs. intra-instance graphs.}
  \label{fig:edge_deletion_remaining}
\end{figure}

In \cref{sec:defederation}, we focused on edge deletion prediction by using the same method than for edge prediction, with limited results.
A natural question is thus: is the poor performance due to noise in the deletion process? To answer this question, we must assess whether edge deletion is a random process (e.g., caused by network instability) or whether it persists over time. In particular, \cref{fig:temp} does not exhibit clear trends and tends to suggest that deletions could be accidental. We test this hypothesis by comparing the edges and nodes deleted between two consecutive weeks (as reported in \cref{fig:temp}) to those that remain deleted over longer time horizons: two weeks, four weeks, and until the end of the measurement period.

We report the results for node deletion in \cref{fig:node_deletion}, and for edge deletion in \cref{fig:edge_deletion} and \cref{fig:edge_deletion_remaining}, depending on the type of graph. The results show that most variations are permanent, at least within the time scale of our study, as soon as they persist beyond two weeks.

The majority of deletions appear to be permanent, except in BookWyrm, which can be explained by the small size of the network, and in Mastodon, where node and edge deletion durations follow a continuum.  
The only other case showing a progressive increase in the proportion of affected edges with respect to duration is Lemmy cross- and intra-instance, which can be explained by the construction of these graphs: they depend on recent message exchanges between instances and may naturally include servers whose users interact only occasionally.

In these cases, edge deletion does not appear to be an artifact of the crawling process, but rather reflects actual communication dynamics in the graph.
 
We also present a baseline for node deletion. Similarly to the case of edges, we report the number of actual deletions among the top-$50$ nodes with the highest scores. We propose to predict deletion for nodes that are the least central in the graph, as this may indicate lower activity. Other metrics, and incorporating additional information about the nodes, could certainly improve these predictions. We report the results in \cref{tab:nodedel}.

\begin{table}
    \centering
    \begin{tabular}{lcccc}
         \toprule
         Graph & Betweenness & Eigenvector centrality & Pagerank & Random\\
         \midrule
         Misskey AU & $6$ & $16$ & $16$ & $7 \pm 2.4$\\
         Misskey fed. & $7$ & $12$ & $11$ & $7 \pm 2.3$\\
         Friendica fed. & $11$ & $13$ & $13$ & $7 \pm 2.4$\\
         Pleroma fed & $13$ & $3$ & $12$ & $8 \pm 2.5$\\
         \bottomrule
    \end{tabular}
    \caption{Comparison of Betweenness, eigenvector centrality and Pagerank scores on different graphs by reporting the number of correct predictions in Top-$50$ scores (the higher the better)}
    \label{tab:nodedel}
\end{table}

\section{Additional remarks about the crawler}
We discuss in this section the robustness of the crawling, and legal procedure prior the crawler deployment. We refer to \cref{subsec:ethical-concerns} for the ethical concerns related to the crawling procedure.

\paragraph{Robustness}
Fig. \ref{fig:evonode} represents the variations of the number of nodes in different Fediverse software.
For all these social networks, the variations looks erratic. Such patterns could raise concerns regarding the robustness of the crawling procedure.
However, these erratic variations are a natural phenomenon of the Fediverse and are not an artifact created by the crawler.

\begin{figure}
    \centering
    \centering
    \includegraphics[width=0.45\linewidth]{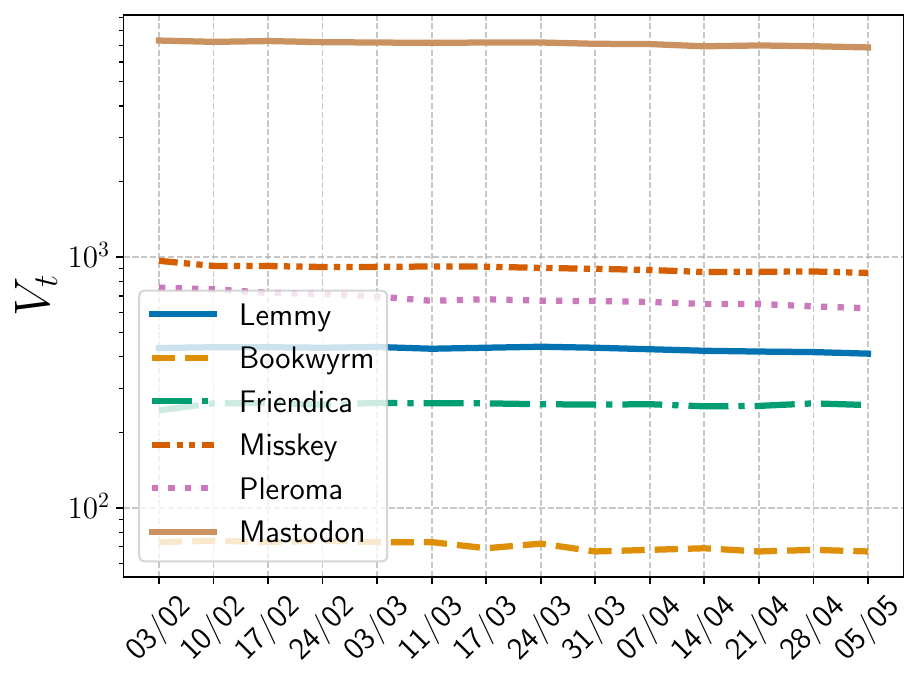}
    \caption{Evolution of the number of nodes over time for the six software of Fedivertex}
    \label{fig:evonode_absolute}
\end{figure}

To put these variations into perspective, \cref{fig:evonode_absolute} represents the overall evolution instead of the node variation between two snapshots. We observe that the number of nodes is relatively stable.
The patterns from \cref{fig:evonode} are barely distinguishable on \cref{fig:evonode_absolute} because they concern a minority of nodes.

Even though these patterns involve only a minority of nodes, they are worth examining. Most instances on the Fediverse are maintained by volunteers who cover the operational and maintenance costs themselves. Many are operated by a single individual who may lack the time, financial resources, or motivation to ensure long-term stability. While the biggest instances benefit from a high stability due to their teams of experienced engineers, smaller nodes can quickly appear, disappear, and eventually reappear.

A natural question that follows from this observation is whether our crawler adequately captures the dynamics of unstable instances. Our approach relies on a curated list maintained by Fediverse Observer, which continuously crawls the Fediverse to discover new instances and assess the availability of existing ones. Notably, instances that are temporarily unavailable remain listed, meaning our crawler does not ignore unstable nodes. Thanks to the frequent updates and broad coverage provided by Fediverse Observer, the resulting instance list is both extensive and up-to-date. We thus argue that our crawler offers a realistic and consistent view of the Fediverse.

Our source code is publicly available: \url{https://github.com/MarcT0K/Franck}.
The crawler implementation has remained consistent since the first crawl, with only minor changes introduced to improve logging completeness.  

\paragraph{Legal compliance}
Before crawler deployment, we have been involved in active discussions with the legal departments of our institutions to cover any potential legal issue. We worked in particular on the compliance with data privacy regulations, especially GDPR. As stressed in \cref{subsec:ethical-concerns}, our graphs only contain aggregated information (i.e., server-level information), and we only query public APIs, and thus are less privacy sensitive than existing works (e.g., the Webis Mastodon corpus \cite{wiegmann_mastodon_2024} that compiled 700 million Mastodon posts). The crawling logs are kept only for the time necessary to confirm the crawl success and deleted afterwards.

Our work is among the first research projects on the Fediverse, and the guidelines for research in this field still need to be refined. Existing guidelines tend to focus on centralized mainstream social media, and are thus not well adapted to scenarios where instances are run by distinct legal entities, possibly subject to different regulations and norms. The discussions with the legal departments allowed us to clarify the legal requirements in this context. We shared our experience with the Network of Alternative Social Media Researchers, a transdisciplinary international collective of academics studying the cultures, technologies, and practices of non-corporate social media\footnote{Testimony at \url{https://www.socialmediaalternatives.org/2025/05/07/asm-research-ethics.html}}, to help formalize a dedicated ethical framework for research on alternative social media.

In addition to the discussions with the legal department, we have also coordinated our efforts with the IT department to avoid any disturbance caused by our crawler. Notably, our crawler is formally declared to the security operations center of our university, is assigned specific IP addresses, and is under the surveillance of the IT department. These measures are complementary to the preventive strategy of a \emph{slow} crawl, aiming to minimize disturbances for the instances, as detailed in \cref{subsec:ethical-concerns}. After four months of experiments, we have received no complaints, neither from our university nor from Fediverse instance administrators.

\end{document}